\documentclass{article} 
\usepackage{iclr2025_conference,times}

\usepackage{amsmath,amsfonts,bm}

\def\eqref#1{equation~\ref{#1}}

\def\1{\bm{1}}

\DeclareMathAlphabet{\mathsfit}{\encodingdefault}{\sfdefault}{m}{sl}
\SetMathAlphabet{\mathsfit}{bold}{\encodingdefault}{\sfdefault}{bx}{n}

\usepackage{graphicx}
\usepackage{hyperref}
\usepackage{url}
\usepackage{subcaption}
\usepackage{booktabs}
\usepackage{tikz}
\usetikzlibrary{positioning}

\title{Multimodal graph representation learning for website generation based on visual sketch}

\author{Tung D. Vu$^\dagger$ \\
College of Engineering and Computer Science\\
VinUniversity\\
Hanoi, Vietnam\\
\texttt{21tung.vd@vinuni.edu.vn} \\
\And
Chung Hoang$^\dagger$ \\
Department of Computer Science \\
Hanoi University of Science and Technology \\
Hanoi, Vietnam \\
\texttt{chunghoang0103@gmail.com} \\
\AND
Truong-Son Hy\thanks{Corresponding Author. $^\dagger$ This work is done during Vu and Hoang's internship at University of Alabama at Birmingham under the supervision of Dr. Truong-Son Hy.} \\
Department of Computer Science\\
The University of Alabama at Birmingham\\
Birmingham, Alabama, United States \\
\texttt{thy@uab.edu}
}

\iclrfinalcopy 
\begin{document}

\maketitle

\begin{abstract}
The Design2Code problem, which involves converting digital designs into functional source code, is a significant challenge in software development due to its complexity and time-consuming nature. Traditional approaches often struggle with accurately interpreting the intricate visual details and structural relationships inherent in webpage designs, leading to limitations in automation and efficiency. In this paper, we propose a novel method that leverages multimodal graph representation learning to address these challenges. By integrating both visual and structural information from design sketches, our approach enhances the accuracy and efficiency of code generation, particularly in producing semantically correct and structurally sound HTML code. We present a comprehensive evaluation of our method, demonstrating significant improvements in both accuracy and efficiency compared to existing techniques. Extensive evaluation demonstrates significant improvements of multimodal graph learning over existing techniques, highlighting the potential of our method to revolutionize design-to-code automation.

Code available at \url{https://github.com/HySonLab/Design2Code}.
\end{abstract}

\section{Introduction}
The Design2Code problem, which involves converting UI designs into functional source code, is a pivotal challenge in software development that lies at the intersection of computer vision, natural language processing, and programming. This task is particularly demanding when generating HTML code from webpage designs, as it requires not only the interpretation of visual elements but also an understanding of their spatial arrangements and hierarchical relationships. While significant advances have been made in both vision-language models and code generation models, combining the strengths of these two fields to solve Design2Code remains an unsolved challenge. One of the fundamental difficulties is that current vision-language models, which excel at generating descriptive text from images \citep{hakimov-schlangen-2023-images, iscen2024retrievalenhancedcontrastivevisiontextmodels, Lin_2024_CVPR}, often fall short when tasked with understanding the intricate structural layout of a design. These models, which typically focus on high-level image features, struggle with capturing the precise relationships between elements such as deeply nested blocks, grid layout, spatial alignment of diverse visual components. On the other hand, code generation models \citep{zan2023largelanguagemodelsmeet, sun2024surveyneuralcodeintelligence, lyu2024automaticprogramminglargelanguage}, despite their proficiency in generating syntactically correct code from textual descriptions, often lack the visual comprehension needed to map these descriptions to complex layouts. This results in syntactically correct code, yet often fails to reflect the intended design. The key to overcoming these challenges lies in developing a model architecture that not only captures the global features of an image but also pays close attention to these critical attributes and relationship, ensuring both breadth and depth in its analysis.

Given these motivations, we introduce a novel framework for the Design2Code problem that integrates multimodal information from webpage screenshots and component-level relationships. Using Optical Character Recognition (OCR) and segmentation models, our approach extracts textual and visual components, organizing them into a multimodal graph that captures the webpage’s structure and serves as a blueprint for generating HTML code. By isolating textual components early through OCR, we enable the segmentation model to more accurately focus on visual components without interference from overlapping text. The multimodal graph maps both textual and visual components, capturing their spatial and semantic relationships. Textual elements are linked semantically, while visual components are connected based on spatial proximity, forming the foundation for code generation. Our model integrates the visual elements and structural relationships within a graph representation, using a Vision-Language Model (VLM) to encode the webpage’s appearance. This combined multimodal data ensures both visual and structural details contribute to generating accurate HTML code. The contributions of this paper are threefold:
\begin{itemize}
    \item We propose a Graph-enhanced Vision-Language Model that seamlessly integrates graph modality, vision modality, and text modality to generate accurate HTML code.
    \item We propose a novel approach to construct graph representations of user interfaces, enabling the model to learn latent structures that effectively bridge visual layouts and code semantics, thereby enhancing UI-to-code translation.
    \item We perform extensive experiments to demonstrate that our approach significantly improves both content accuracy and layout fidelity compared to baseline models, offering a robust solution to the Design2Code problem.
\end{itemize}

In the following sections, we first explore the existing research on the problem of converting sketches into website source code in Section \ref{sec:related}. Section \ref{sec:method} then details the methodology employed in this study. The experimental results are discussed in Section \ref{sec:experiment}, followed by conclusions, limitations, and future research in Section \ref{sec:conclusion} and Section \ref{sec:limitation}, respectively.
\section{Related work} \label{sec:related}

\subsection{Design2Code}
\paragraph{Approaches to Automating Design2Code Generation.}

The Design2Code (or UI2Code) problem has gained significant attention as researchers work to automate the conversion of user interface (UI) designs into source code. Early methods primarily relied on rule-based systems that assigned UI components to predefined code templates \citep{Nguyen2015ReverseEM}, offering limited flexibility. As the complexity of web design increased, newer approaches adopted deep learning techniques for enhanced automation and accuracy. Notable models like Pix2Code \citep{10.1145/3220134.3220135} and Sketch2Code \citep{robinson2019sketch2codegeneratingwebsitepaper} utilize convolutional neural networks (CNNs) to transform UI screenshots into single-file code, with Pix2Code demonstrating cross-platform versatility and Sketch2Code focusing on wireframe sketches. Recent efforts, such as those by \cite{soselia2023learninguitocodereversegenerator}, treat image-to-code generation as a Reinforcement Learning problem, using the Intersection over Union (IoU) score as a reward signal. Evaluations on benchmarks such as Design2Code \cite{si2024design2codefarautomatingfrontend} have utilized open-access models from the GPT, Claude, and Gemini families, achieving strong performance without the need for fine-tuning. \cite{GUICG2023} adopts a hybrid approach that leverages computer vision techniques to better capture the structure and components of UI images, in combination with deep learning models for code generation. Other works draw inspiration from the spatial or hierarchical organization of UI components to guide the model or adjust the inference strategy \citep{declareUI, gui2025uicopilot}. The work most similar to ours is \cite{graph4gui}, which also employs computer vision techniques and rule-based methods to construct a graph, followed by training a Graph Neural Network to understand the high-level structure of UI components. However, their primary objective is not code generation. In contrast, our work enhances Visual-Language Models (VLMs) of suitable size by incorporating graph learning, aiming to better capture structural features and improve accuracy in the Design2Code task.

\paragraph{Datasets for Design2Code.} 
A crucial factor in advancing Design2Code methods is the availability of suitable datasets and benchmarks. One foundational dataset in this field is Pix2Code, an open-source, synthesized collection of 1,750 UI screenshots paired with corresponding source code. The Sketch2Code dataset \citep{shantamvijayputra-2022} builds on this by converting Pix2Code screenshots into hand-drawn wireframe representations, adding an additional layer of abstraction for model training. Additionally, the Websight dataset from Hugging Face \citep{Hugo2024} offers a larger scale, comprising 2 million triplets of HTML code, screenshots, and generated descriptions. While these synthesized datasets are critical for enabling large-scale training, they often lack the diversity and complexity found in real-world web pages. Similarly, Vision2UI \cite{Gui2024VISION2UIAR} provides 20,000 samples extracted from real-world scenarios through a meticulous process of data collection, cleaning, and filtering. Although these datasets advance the field by introducing more complex and diverse data, the WebUI dataset \citep{10.1145/3544548.3581158}, although comprehensive and rich in metadata, lacks the corresponding source code and is therefore not used in Design2Code tasks. To bridge the gap between synthesized datasets and real-world applications, the Design2Code benchmark \citep{si2024design2codefarautomatingfrontend} was introduced as the first dataset of real-world web pages specifically for evaluating design-to-code models. Our work leverages the available WebSight datasets to train and benchmark the effectiveness of graph learning in the Design2Code task. 

\paragraph{Existing Benchmarks.} htmlBLEU \citep{soselia2023learninguitocodereversegenerator} uses DOM-tree matching between text inputs and enhances this method by incorporating attribute matching and assigning additional weights to HTML keywords in the BLEU calculation. Similarly, \citep{Gui2024VISION2UIAR} introduces TreeBLEU, an improvement over BLEU, which evaluates the match between generated HTML by comparing 1-height subtrees of the DOM from both the hypothesis and reference structures, enabling a more accurate comparison. Web2Code \citep{Yun2024Web2CodeAL} proposes metrics based on large language models (LLMs) to evaluate webpage understanding and code generation; however, resource constraints limit the real-time application of this method in development. Other vision-related metrics in \cite{soselia2023learninguitocodereversegenerator} have used simpler metrics such as Mean Squared Error (MSE) or Structural Similarity Index (SSIM) based on pixel values. Our work utilizes various existing evaluation metrics, excluding VLM prompting methods, to provide a comprehensive assessment of model performance in the Design2Code domain, building on previous approaches that integrated both VLMs and LLMs for evaluation.

\subsection{From Multimodal Input To Text With Vision Language Models}

Vision-Language Models (VLMs) have transformed how machines interpret and generate language from visual inputs, enabling a range of multimodal tasks. Early models like VisualBERT \citep{li2019visualbertsimpleperformantbaseline} and LXMERT \citep{tan-bansal-2019-lxmert} laid the groundwork by combining BERT \citep{devlin2019bertpretrainingdeepbidirectional} with visual encoders, using a dual-stream approach that limited deep fusion of visual and textual information. Later models, such as UNITER \citep{10.1007/978-3-030-58577-8_7} and VilBERT \citep{li2019visualbertsimpleperformantbaseline}, achieved tighter integration and better cross-modal interactions but struggled with tasks requiring complex reasoning, like code generation. DeepMind’s Flamingo \citep{alayrac2022flamingo} excels at processing interleaved visual and textual data, performing well in few-shot learning tasks. Its open-source counterpart, OpenFlamingo \citep{awadalla2023openflamingoopensourceframeworktraining}, and models like BLIP \citep{pmlr-v162-li22n}, ALBEF \citep{NEURIPS2021_50525975}, Idefics \citep{NEURIPS2023_e2cfb719}, and CogVLM \citep{wang2024cogvlmvisualexpertpretrained} enhance feature alignment, improving text generation from visual inputs.

\subsection{Graph Neural Networks}
Among the most prominent variants of Graph Neural Networks (GNNs) \citep{kipf2017semisupervised, NEURIPS2019_103303dd, 9239975, NEURIPS2022_5d4834a1}, Graph Convolutional Networks (GCNs) can still stand out due to their ability to generalize convolution to graph-structured data, leveraging the connectivity of nodes to aggregate information from their neighbors. By utilizing a recursive neighborhood aggregation process, often referred to as "message passing," GCNs enable each node to update its representation by integrating its own features with those of its neighbors. This allows the model to capture both local node information and the broader structural context in a scalable and efficient manner. Their simplicity, efficiency, and strong theoretical foundation make GCNs highly adaptable to a wide range of graph-based tasks.
Formally, in GCNs, the forward pass for each layer can be describe as:
$$
    H^{(l+1)} = \sigma (\tilde{D}^{-\frac{1}{2}}\tilde{A}\tilde{D}^{-\frac{1}{2}}H^{(l)}W^{(l)}).
$$
Here, $\tilde{A} = A + I_{N}$ denotes the adjacency matrix of the undirected graph $\mathcal{G}$ with inserted self-loops, $I_{N}$ is the identity matrix, and $\tilde{D}_{ii} = \sum_{j} \tilde{A}_{ij}$ is its diagonal degree matrix. $W^{(l)}$ is a layer-specific trainable weight matrix. $\sigma(\cdot)$ denotes a non-linear activation function such as $\mathbf{ReLU}(\cdot)$. $H^{(l)} \in \mathbb{R}^{N \times D}$ is the matrix of activation in the $l^{th}$ layer, with $H^{(0)} = X$ is the feature matrix. 
In GCNs, the ``convolution'' is applied to nodes in a graph, where the neighbors of each node are analogous to the local region in a CNN. The key idea is that each node aggregates information from its neighbors based on the graph’s connectivity. Instead of spatially local filters, GCNs use the graph adjacency matrix to propagate and aggregate features from a node’s neighborhood. This process updates each node’s representation by combining its features with those of its neighbors in a way similar to how convolution combines information from nearby pixels in CNNs. Its node-wise formulation is given by:
$$
    x_{i}^{'} = \theta^{T} \sum_{j \in \mathcal{N}(i) \bigcup \{i\}} \frac{e_{j,i}}{\sqrt{\hat{d}_{j} \hat{d}_{i}}} x_{j},
$$
in which $\hat{d}_{i} = 1 + \sum_{j \in \mathcal{N}(i)} e_{j, i}$ denotes in-degree of node $i$ and $e_{j, i}$ denotes the edge weight from source node $j$ to target node $i$.
By drawing on these related works, our approach integrates the strengths of vision-language models, code generation techniques, and graph representation learning. 

\section{Methodology} \label{sec:method}

\begin{figure*}[ht]
    \centering
    \includegraphics[width=\linewidth]{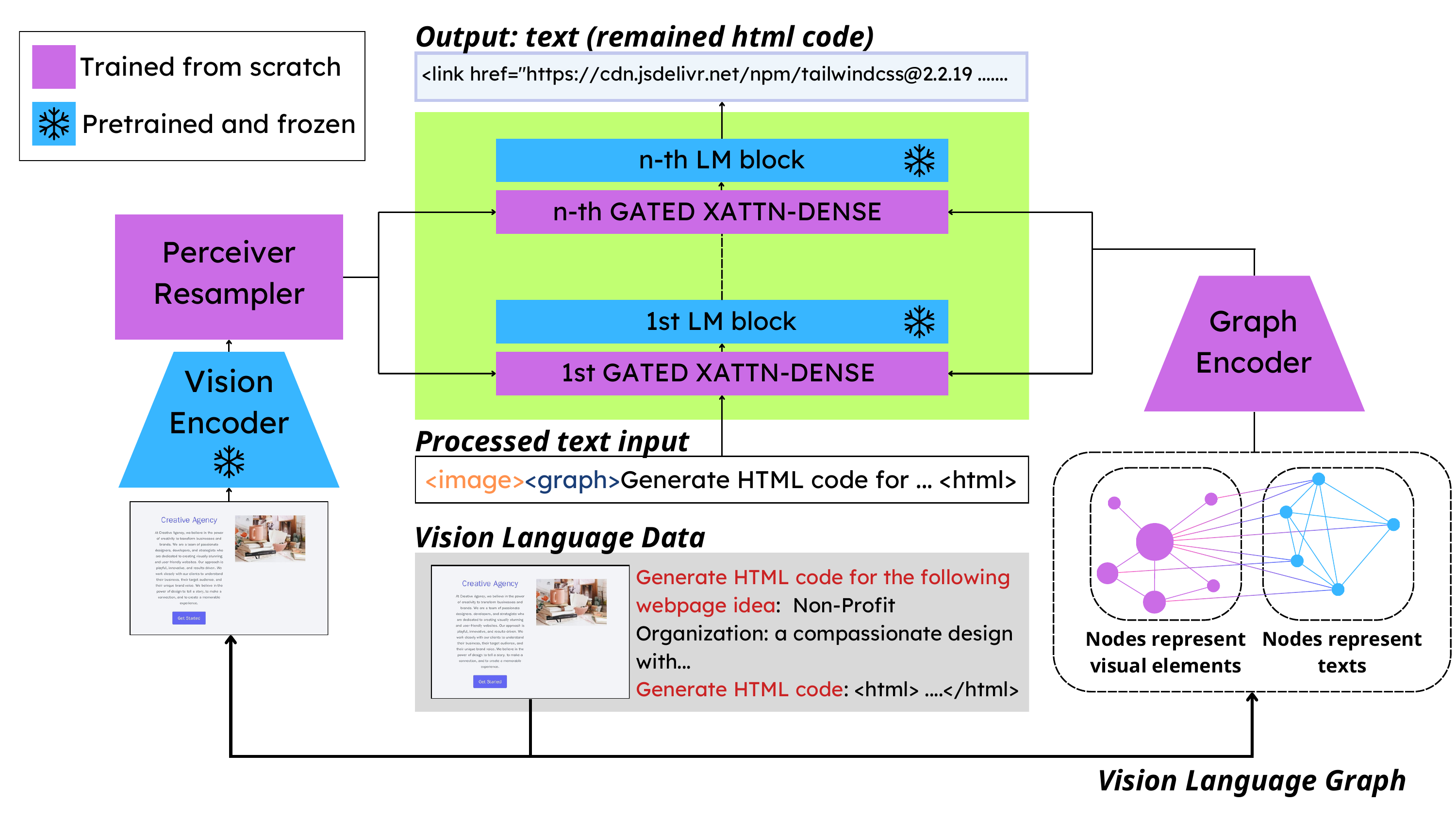}
    \caption{Overview of the Graph-Enhanced Multimodal Architecture for Generating HTML Code from Visual Sketches. The architecture integrates visual and structural information through a Vision Encoder and a Graph Encoder, both of which condition the language model using GATED XATTN-DENSE blocks—our Cross-Attention mechanism 
    for multimodal conditioning}
    \label{fig:model}
\end{figure*}

    In this work, we propose a novel framework to address the Design2Code problem, which aims to automatically generate webpage's code from visual designs. Our approach leverages both an Optical Character Recognition (OCR) model and a Segmentation model to identify and extract the individual components from a given webpage screenshot. These extracted components, along with their spatial features, are then used to construct a graph representing the webpage structure. To further enhance the quality of code generation, we introduce a graph-enhanced vision language model that integrates visual and structural information to produce both the content and the corresponding HTML code for the webpage.


\begin{figure}[htbp]
  \centering
  \begin{minipage}[b]{0.42\linewidth}
    \centering
    \includegraphics[width=\linewidth]{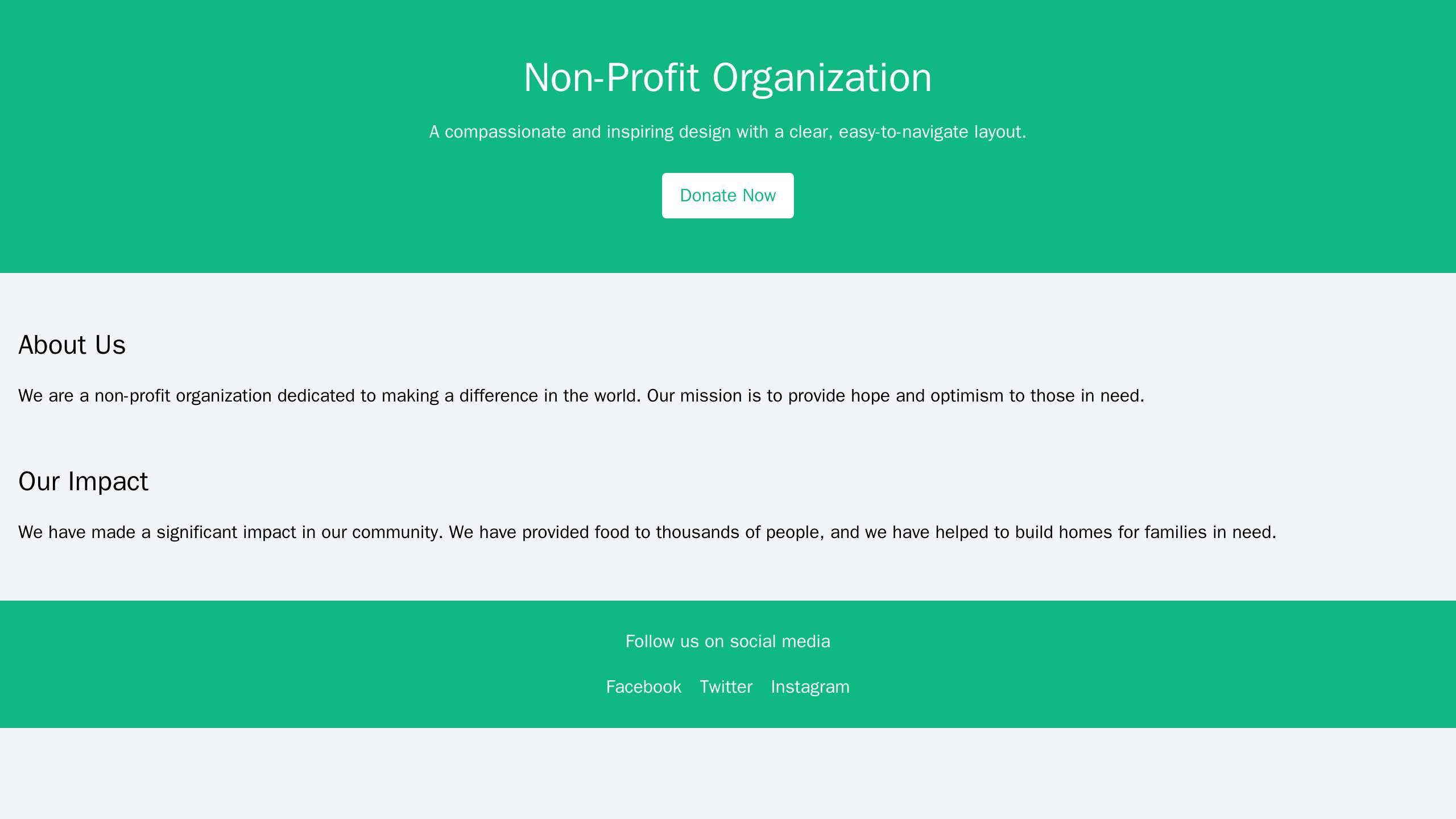}
    \subcaption{Original screenshot}
  \end{minipage}%
  \hspace{0.02\linewidth}
  \begin{minipage}[b]{0.42\linewidth}
    \centering
    \includegraphics[width=\linewidth]{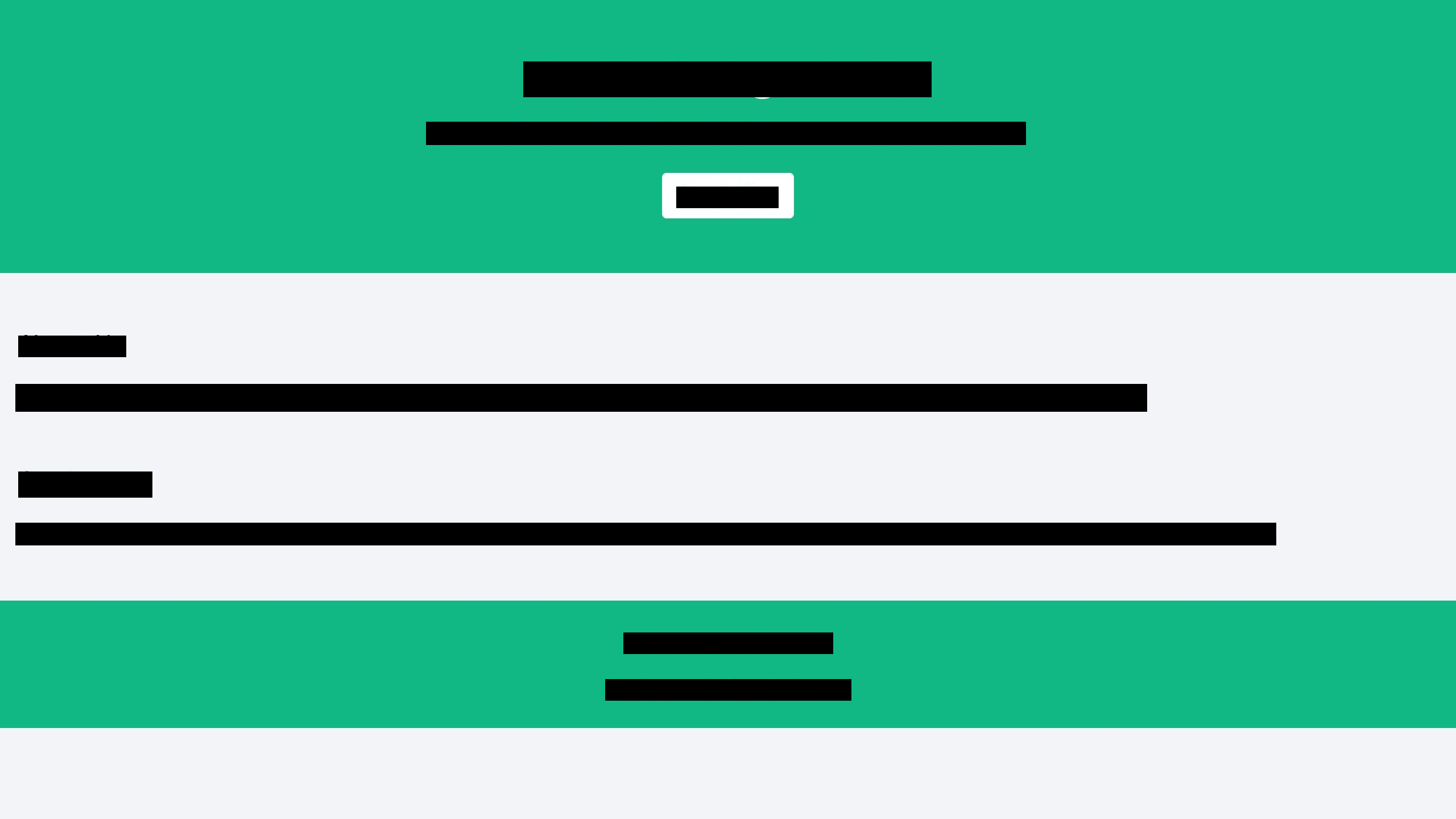}
    \subcaption{Text‐masked by PaddleOCR}
  \end{minipage}%
  \hspace{0.02\linewidth}
  \begin{minipage}[b]{0.42\linewidth}
    \centering
    \includegraphics[width=\linewidth]{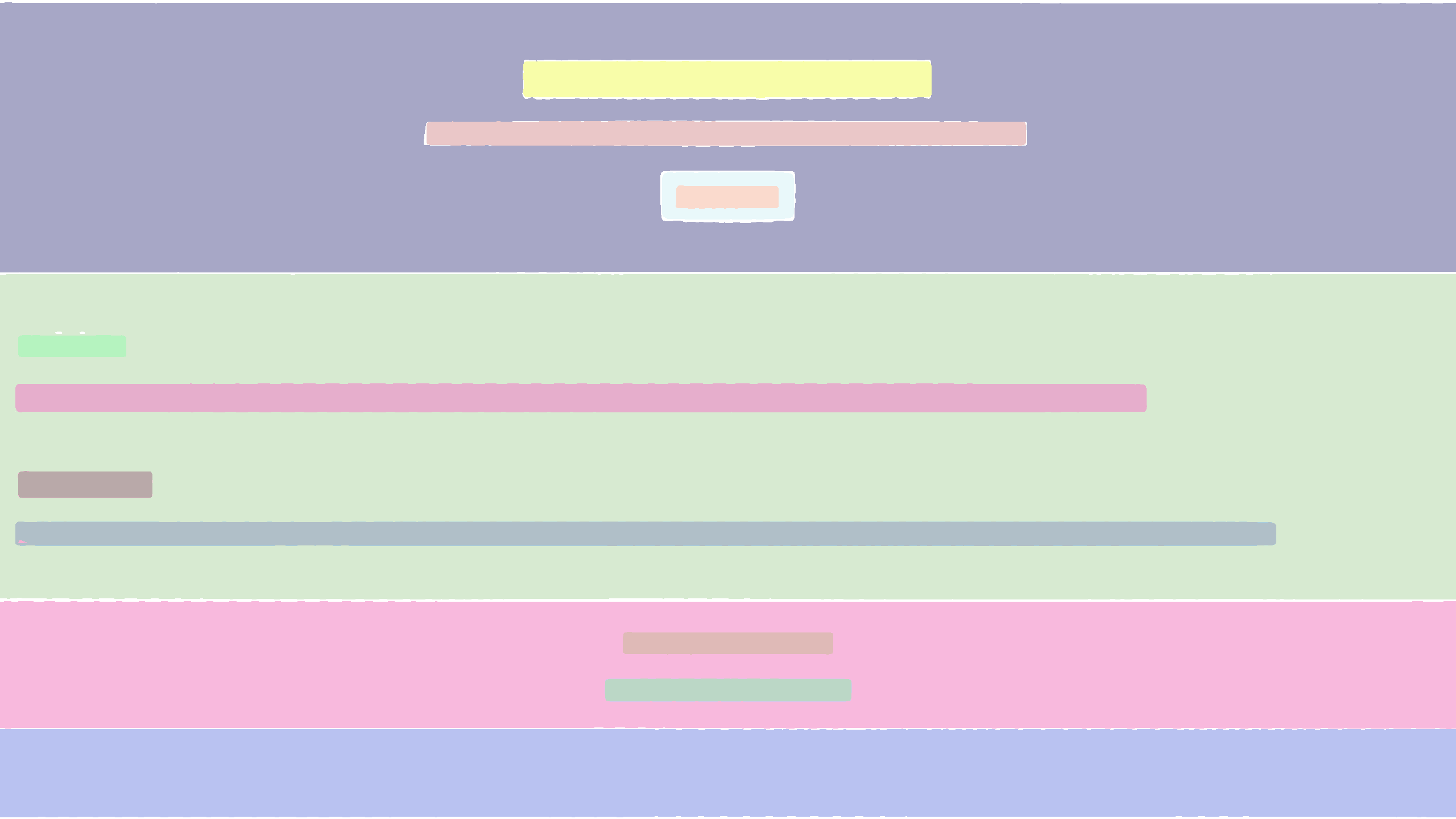}
    \subcaption{SAM segmentation}
  \end{minipage}

  \caption{Illustration of the component‐extraction pipeline.}
  \label{fig:component_extraction}
\end{figure}

    Webpage screenshots often present significant challenges due to the complexity and variety of their content. These images typically contain noisy visual elements, such as overlapping text, dense layouts, and intricate designs, which make it difficult for traditional segmentation models to accurately identify and isolate individual components. To address this issue, we utilize a two-step approach combining Optical Character Recognition (OCR) and the Segment Anything Model (SAM) \footnote{https://github.com/facebookresearch/segment-anything}.
    \paragraph{Text Extraction Using OCR.}
    We begin by applying an OCR model to detect and extract all textual content from the webpage. The OCR model effectively identifies text components, such as headers, paragraphs, and embedded labels, which are often misinterpreted or missed by a segmentation model working directly on the raw screenshot. Once the OCR has successfully extracted these textual components, we remove them from the image by replacing them with black blocks. This step reduces the visual complexity of the image, allowing the segmentation model to focus on non-textual elements.
    \paragraph{Segmentation of Non-Textual Components.}
    With the text removed and masked by black blocks, we apply the SAM to the modified screenshot to extract the remaining components, such as images, buttons, and containers. The black blocks act as placeholders, guiding the segmentation model to recognize distinct visual elements without being confused by text. This process ensures that the non-text components are segmented more accurately and that their relationships with the surrounding elements are better preserved.
    \paragraph{Final Components Combination.}
    After segmentation, we merge the OCR-extracted textual content with the corresponding visual components from the original webpage. The final set of components includes both the textual content identified by the PaddleOCR \footnote{https://github.com/PaddlePaddle/PaddleOCR} model and the rectangular blocks representing the elements extracted by the segmentation model. These components are then used to construct a multimodal graph that represents the content, structure and layout of the webpage.
    

    By separating the handling of textual and visual components, our approach significantly improves the precision of component extraction and ensures that both types of content are correctly interpreted for the subsequent stages of the framework.

    \subsection{Graph Construction}
    \label{sec:graph_construction}
    To capture both structural and semantic relationships among webpage components, we construct a multimodal graph where each node represents a component extracted during the component extraction phase. These components include textual elements and visual blocks. The graph is designed to encode both content-based and spatial dependencies, enabling the model to effectively reason over layout and semantics during code generation.
    \paragraph{Node Representation.}
    Each node in the graph corresponds to a distinct component from the webpage screenshot. Textual nodes (e.g., headers, paragraphs, and footers) are extracted via the OCR pipeline, while visual nodes (e.g., images, buttons, logos, and containers) are obtained through the segmentation model. By incorporating both textual and visual nodes, the graph captures the diverse content and spatial structure inherent in webpage designs.
    \paragraph{Edge Construction.}
    Edges are defined to reflect meaningful relationships between components, capturing both semantical associations and spatial proximity. We construct three types of edges:
    
    \begin{itemize}
        \item \textbf{Textual-to-textual edges}: All textual nodes are fully connected, forming a complete subgraph. This design captures the semantic coherence among textual components, enabling the model to infer relationships such as heading-body associations or grouped lists. Fully connecting these nodes allows for richer interactions across the textual content.
        
        \item \textbf{Visual-to-visual edges}: Edges between visual nodes are added based on spatial overlap. Specifically, an edge is established if the intersection-over-union (IoU) between two visual components exceeds 80\%. This criterion connects elements that are physically close or overlapping in the layout, allowing the model to learn local visual grouping and alignment patterns.
    
        \item \textbf{Textual-to-visual edges}: Similar to the visual edges, a connection is added between a textual node and a visual node if their bounding boxes intersect with an IoU greater than 80\%. This links labels to associated elements (e.g., a button’s caption or an image’s alt-text), preserving the multimodal alignment between content and design.
    \end{itemize}

    \subsection{Graph-Enhanced Vision Language Model} 
    \label{sec:model}
    We design a graph-enhanced vision-language model to generate accurate webpage content and HTML code by conditioning a language model on both visual and graph-based information. This architecture combines the strengths of graph neural networks (GNNs), vision encoders, and pretrained language models, creating a multimodal system that captures the structure, layout, and content of the webpage.
    \paragraph{Graph Convolutional Network for Graph Embeddings.} 
    We employ a graph convolutional network (GCN) to encode the multimodal graph constructed from the webpage components. The GCN serves as the graph encoder, learning informative embeddings for each node in the graph. At the Layer 0 of the GCN, we initialize the node embeddings using the features extracted from each textual and visual component. For the feature extraction, we utilize CLIP \citep{pmlr-v139-radford21a}, a powerful multimodal vision-language model. CLIP allows us to obtain rich, contextualized embeddings for visual and textual components. These embeddings are passed into the GCN, where subsequent layers refine the node embeddings by considering the webpage's structure and sematic relationships between components. After finetuning, the GCN effectively learns high-level representations that capture the relationships between different webpage components, ensuring that both semantic relevance and spatial relationships are encoded.
    \paragraph{Vision Encoder with Perceiver Resampler.}
    Vision Encoder with Perceiver Resampler
    To encode the visual information from the entire webpage screenshot, we draw inspiration from the Flamingo model and incorporate a Vision Encoder alongside Perceiver Resampler layers. The Vision Encoder processes the full webpage image and extracts high-level visual features that provide contextual understanding of the page layout, color schemes, and design elements. The Perceiver Resampler layers \citep{perceiver} are applied to produce a fix small numbers of visual tokens from a high-resolution webpage screenshot.
    \paragraph{Cross-Attention for Multimodal Conditioning.}
    The core of our model lies in how we combine the graph modality and vision modality to condition the language model for text and code generation. We use cross-attention layers, which are interleaved between the pretrained layers of a standard language model, to fuse the multimodal inputs into the token prediction process. Specifically, freshly initialized cross-attention layers are introduced between the existing layers of the pretrained language model. These layers take the embeddings from both the vision modality and the graph modality, allowing the language model to incorporate multimodal information at every step of the token generation process. This approach is more aggressive than simple concatenation or late fusion, as it enables a deeper integration of visual and graph information throughout the entire language modeling process. The process is formalized as follows:
    \begin{displaymath}
        \mathbf{E}_{\text{L}}^i = \text{GCA}(\mathbf{X},  \mathbf{E}_\text{L}^{i-1}) +  \text{GCA}(\mathbf{Z}, \mathbf{E}_\text{L}^{i-1})  + \mathbf{E}_\text{L}^{i-1},
    \end{displaymath}
    where $\mathbf{X}$ represents the vision embedding, $\mathbf{Z}$ represents the graph embedding, and $\mathbf{E}_{\text{L}}^i$ is the output at the $i^{th}$ model layers, which integrates both graph and visual information into the textual embedding. The operator $\text{GCA}(\mathbf{A}, \mathbf{B})$ denotes the Gated-Cross Attention mechanism, which performs attention between two embeddings $\mathbf{A}$ and $\mathbf{B}$ from different modalities. 

    \begin{figure*}[ht]
        \centering
        \includegraphics[width=0.8\linewidth]{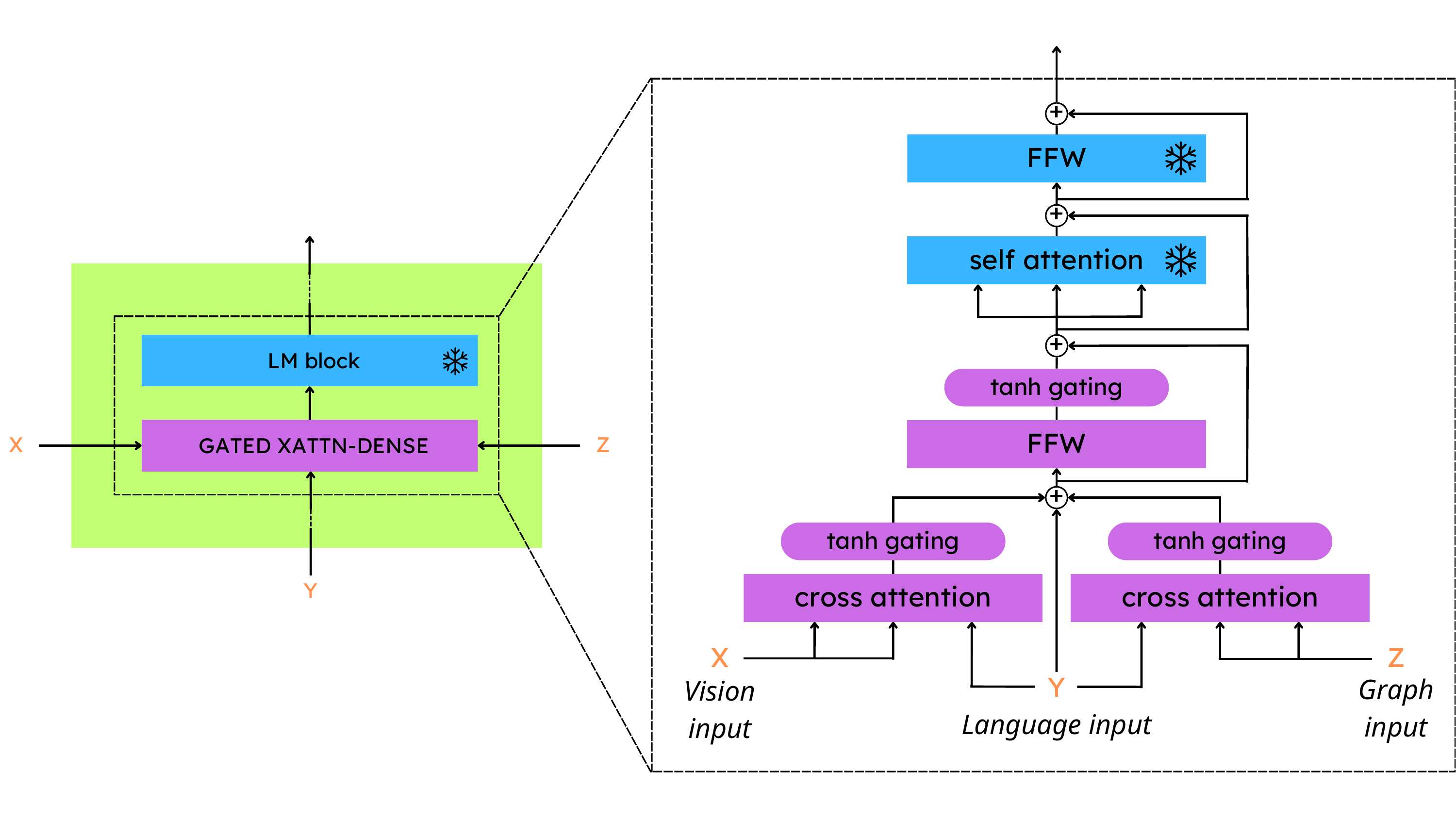}
        \caption{Gated Cross Attention Block. The Gated Cross Attention Block integrates three input modalities - vision ($X$), language ($Y$), and and graph ($Z$) through cross-attention layers, followed by tanh gating to control information flow.}
        \label{fig:model}
    \end{figure*}
    
    The cross-attention mechanism ensures that the model can attend to both the full-page screenshot and the relationships between webpage components when generating the next token. By doing so, the model not only generates accurate content, but also respects the spatial and semantic coherence of the original webpage design.

\section{Experiment} \label{sec:experiment}

\subsection{Experimental Setup}
\begin{table}[ht]
\centering
\begin{tabular}{l|cccc}
\toprule
\textbf{}           & \textbf{WebSight v0.1} & \textbf{WebSight v0.2} & \textbf{Design2Code} & \multicolumn{1}{c}{\textbf{WebSight benchmark}} \\ \midrule
Size                & 823K                      & 1.92M               & 484                     & 500                         \\ 
Purpose             & Training & Training & Testing & Testing \\ 
Avg Length (tokens) & 647±216                   & 708±265            & 31216±23902             & 723±271                     \\ 
Avg Tag Count       & 19±8                      & 19±7         & 158±100                 & 20±7                  \\ 
Avg DOM Depth       & 5±1                       & 6±1          & 13±5                    & 6±1                  \\ 
Avg Unique Tags     & 10±3                      & 11±3         & 22±6                    & 11±3                 \\ \bottomrule
\end{tabular}
\caption{\label{table:data_complexity} Comparison of dataset statistics between WebSight v0.1, WebSight v0.2, Design2Code benchmark, and collected samples from WebSight v0.2 as benchmark.}
\end{table}

We examined both real-world and synthetic data sources. Real-world datasets provide authentic UI examples but pose challenges such as instability during training and complex structures. Due to these issues and limited computing resources, we focus on synthetic data. We trained our model on a curated subset of the Websight Hugging Face dataset, which contains 20,000 HTML code and screenshot pairs. This subset was chosen for its diversity and manageable size, avoiding the noise and instability of real-world data. Another subset of Websight Hugging Face was used as a benchmark for comparison.

Table \ref{table:data_complexity} presents a comparison of dataset statistics between WebSight v0.1, WebSight v0.2, Design2Code, and a WebSight benchmark subset. The data highlights key metrics such as average length, tag count, DOM depth, and unique tags. The WebSight benchmark subset (500 samples) shows average values for length (723±271 tokens), DOM depth (6±1), and tag count (20±7), which closely resemble the overall characteristics of WebSight v0.2. This alignment suggests that the selected WebSight benchmark subset appropriately represents the complexity of the full WebSight dataset, making it a suitable choice for evaluating models alongside datasets like Design2Code.

In terms of evaluation, we employed several metrics to assess performance. These include Block-match, which evaluates the spatial alignment of design blocks; Text, focusing on textual accuracy; and Position, Color, and CLIP, which assess visual and layout fidelity. Additionally, we considered traditional metrics such as MSE (Mean Squared Error) and SSIM (Structural Similarity Index) for image quality, as well as more specialized metrics like TreeBLEU and htmlBLEU for assessing code similarity and structure.

\subsection{Quantitative Results}

\begin{table}[ht]
\centering
\begin{tabular}{lccccc}
\toprule
\\[-1em]
\textbf{}           & \textbf{Block-Match \(\uparrow\)} & \textbf{Text \(\uparrow\)}  & \textbf{Position \(\uparrow\)} & \textbf{Color \(\uparrow\)}  & \textbf{CLIP \(\uparrow\)}   \\ \midrule
\\[-1em]
\multicolumn{6}{c}{\textbf{Websight HF benchmark}} \\ \midrule
\\[-1em]
OURS-graph          & 24.94                & 79.33          & 70.52             & 75.41           & \textbf{91.36}  \\ 
\\[-1em]
OURS-no-graph       & 21.60                & 76.06          & 66.21             & 69.70           & 89.63           \\ 
\\[-1em]
Gemini-prompting    & \textbf{98.21}       & \textbf{99.47} & \textbf{78.29}    & \textbf{83.83}  & 89.85           \\ \midrule
\\[-1em]
\multicolumn{6}{c}{\textbf{Design2Code benchmark}} \\ \midrule
\\[-1em]
OURS-graph          & 2.63                & 47.15           & 37.97             & 40.07           & 82.63           \\ 
\\[-1em]
OURS-no-graph       & 2.51                & 50.15           & 37.37             & 43.30           & 82.90           \\ 
\\[-1em]
Gemini-prompting    & \textbf{90.64}       & \textbf{95.92} & \textbf{78.06}    & \textbf{73.24}  & \textbf{88.37}   \\ 
\\[-1em]
Llava-v1.6-Mistral-7b-hf & 40.77 & 74.26 & 60.27 & 50.86 & 81.33 \\ 
\\[-1em]
Qwen2.5-VL-7B-Instruct    & 2.04  & 44.64 & 37.50 & 41.00 & 80.47 \\ 
\\[-1em]
Idefics3-8B-Llama3    & 1.83 & 39.86 & 31.44 & 36.17 & 81.12 \\ \bottomrule
\end{tabular}
\caption{Performance Comparison on Websight HF and Design2Code Benchmarks. This table presents the performance of OURS-graph, OURS-no-graph, Gemini prompting, and some state-of-the-art open-sourced VLMs on various metrics for Design2Code tasks across two benchmarks: Websight HF and Design2Code. Gemini prompting demonstrates superior performance across most metrics. OURS-graph shows better performance than OURS-no-graph, especially in visual metrics, Block-Match and Position, reflecting the advantage of incorporating graph representation for structural alignment and visual fidelity.}
\end{table}

\begin{table}[ht]
\centering
\begin{tabular}{lccccc}
\toprule
\\[-1em]
\textbf{}           & \textbf{BLEU \(\uparrow\)}  & \textbf{HTML-BLEU \(\uparrow\)}  & \textbf{MSE \(\downarrow\)}    & \textbf{SSIM \(\uparrow\)}  & \textbf{TreeBLEU \(\uparrow\)} \\ \midrule
\\[-1em]
\multicolumn{6}{c}{\textbf{Websight HF benchmark}} \\ \midrule
\\[-1em]
OURS-graph          & \textbf{43.83} & \textbf{52.55}     & \textbf{35.91}  & \textbf{81.16} & \textbf{44.29}    \\ 
\\[-1em]
OURS-no-graph       & 39.69          & 48.19              & 43.20           & 77.57          & 37.21             \\ 
\\[-1em]
Gemini-prompting    & 41.37          & 34.87              & 63.60           & 76.58          & 15.52             \\ 
\\[-1em] 
\midrule
\\[-1em]
\multicolumn{6}{c}{\textbf{Design2Code benchmark}} \\ \midrule
\\[-1em]
OURS-graph          & 2.49           & 7.15               & 67.67           & 65.82          & 16.59              \\ 
\\[-1em]
OURS-no-graph       & 2.03           & 7.27      & 75.69           & 65.14          & 16.71              \\ 
\\[-1em]
Gemini-prompting    & \textbf{6.45}  & 7.17               & \textbf{48.55}  & \textbf{69.72} & \textbf{24.83}      \\ 
\\[-1em]
Llava-v1.6-Mistral-7b-hf & 1.30            & 6.66                & 74.75             & 65.85            & 6.93     \\ 
\\[-1em]
Qwen2.5-VL-7B-Instruct    & 1.70  & \textbf{8.15}               & 77.33  & 58.50 & 9.67  \\ 
\\[-1em]
Idefics3-8B-Llama3    & 0.59  & 5.55               & 72.05  & 59.59 & 4.77   
\\ \bottomrule
\end{tabular}
\caption{Performance on Traditional Metrics for Websight HF and Design2Code Benchmarks. This table compares OURS-graph, OURS-no-graph, Gemini prompting, and some state-of-the-art open-sourced VLMs using traditional metrics - BLEU, HTML-BLEU, MSE, SSIM, and TreeBLEU. On the Websight HF benchmark, OURS-graph outperforms both baselines across all metrics, indicating its effectiveness in generating accurate and visually coherent HTML code. On the Design2Code benchmark, OURS-graph maintains superiority over OURS-no-graph, further demonstrating the value of graph-based representation in improving structural and visual fidelity. Gemini as an API model performs better on most metrics including BLEU, MSE, SSIM, and TrueBLEU.}
\end{table}

\subsection{Qualitative Results}

\begin{table}[ht]
\centering
\caption{Qualitative comparison between graph and w/o graph methods}
\scalebox{1.4}{
\begin{tabular}{ccc}

\includegraphics[width=.2\linewidth]{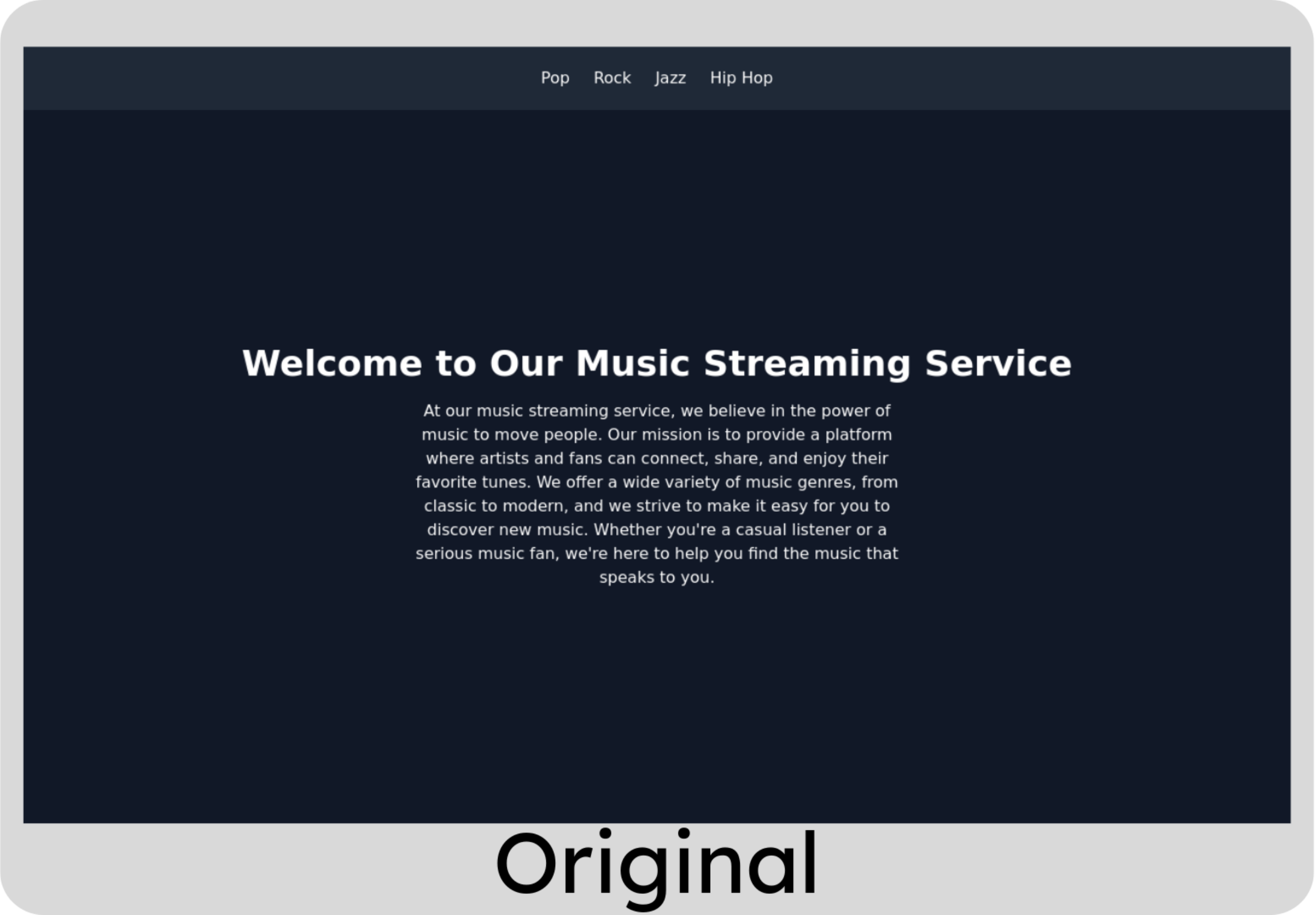} & \includegraphics[width=.2\linewidth]{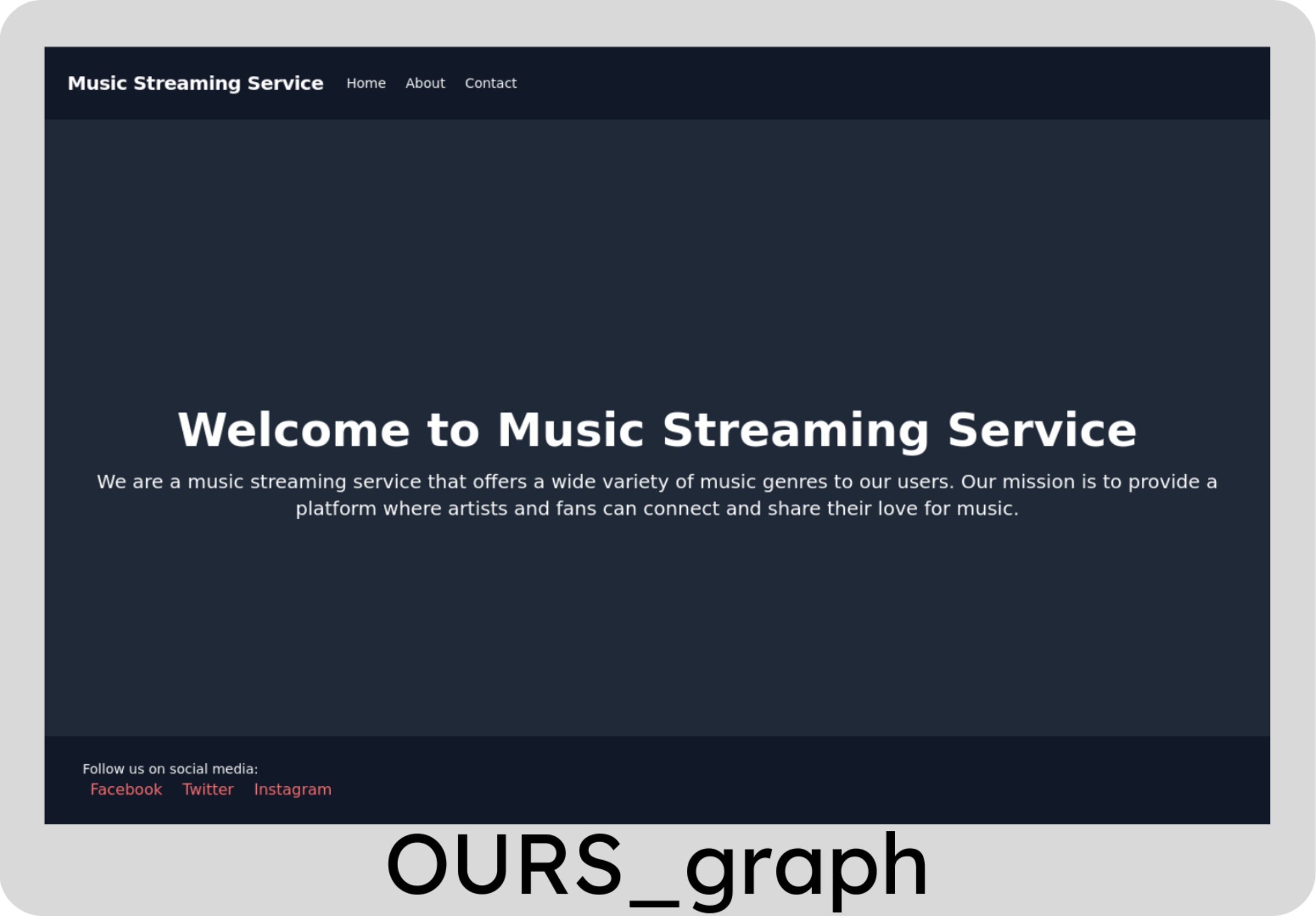} & \includegraphics[width=.2\linewidth]{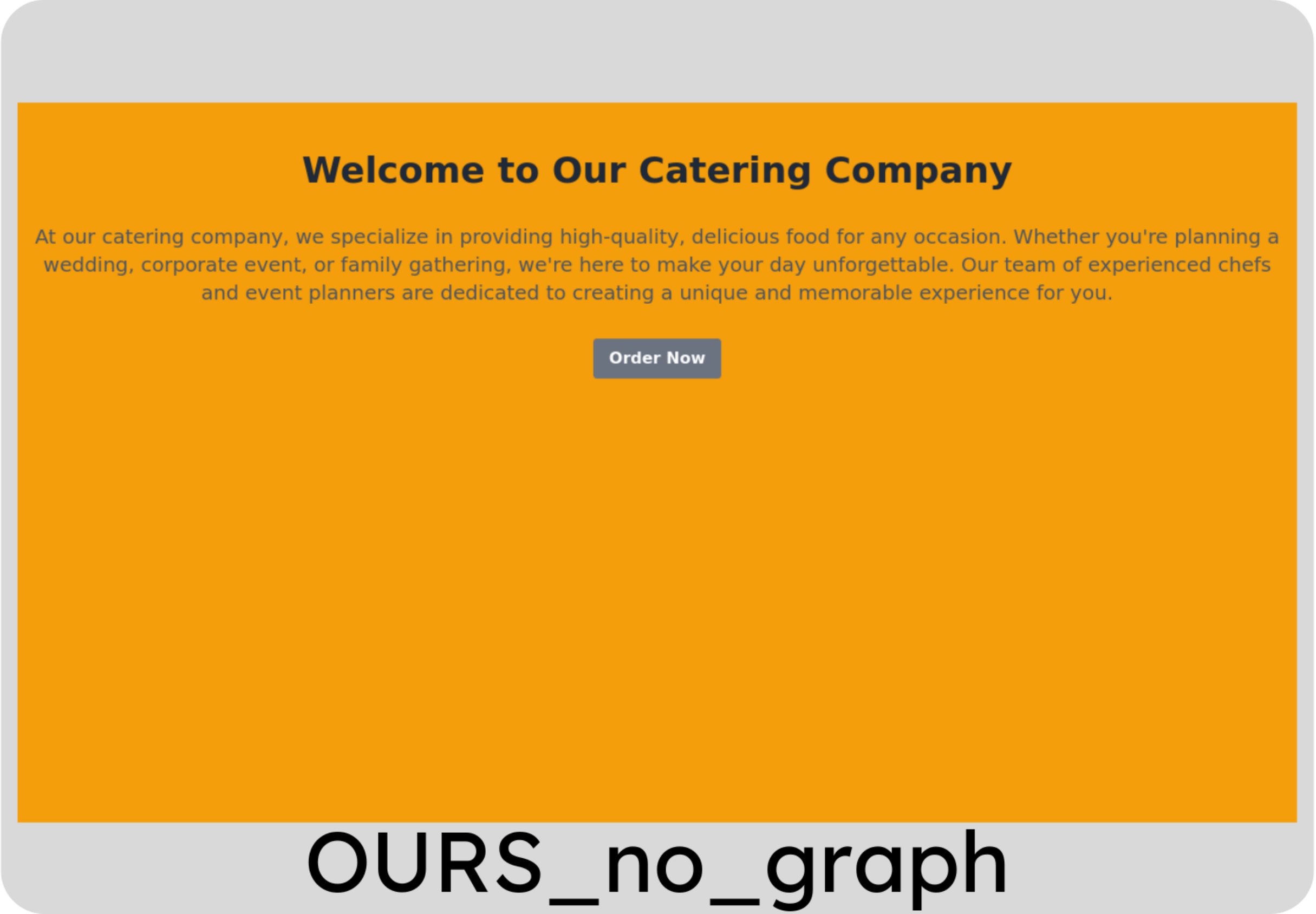} \\

\includegraphics[width=.2\linewidth]{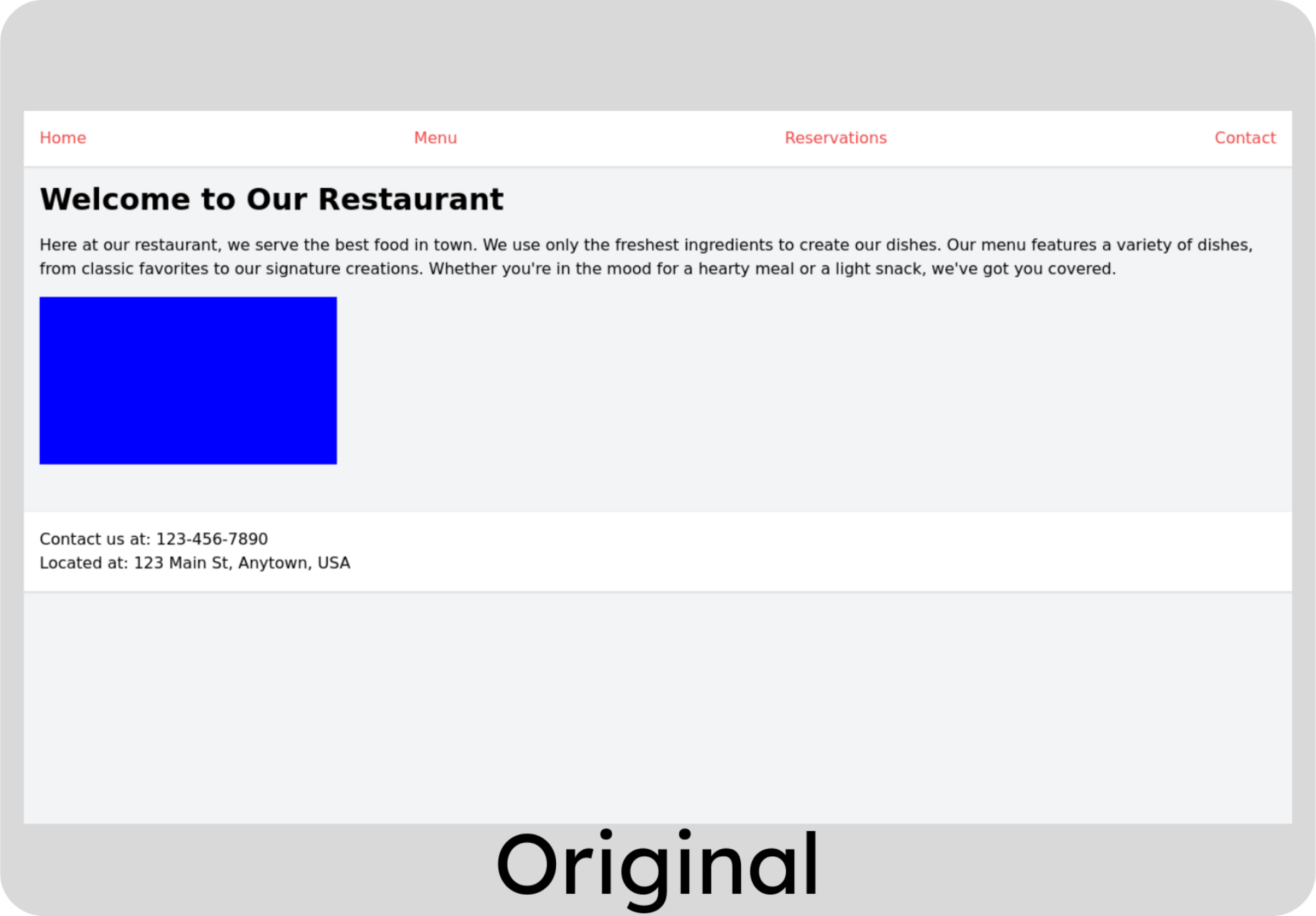} & \includegraphics[width=.2\linewidth]{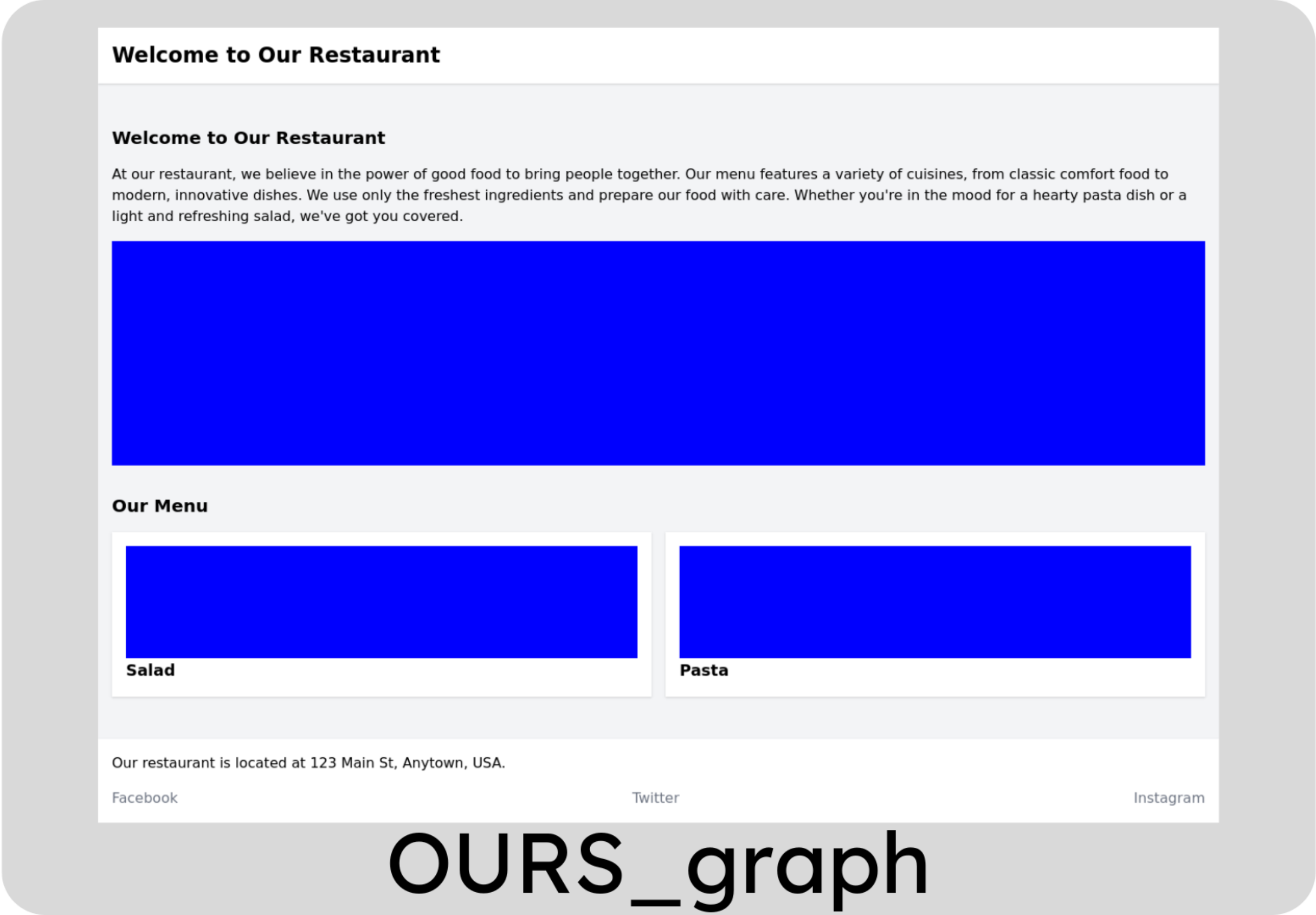} & \includegraphics[width=.2\linewidth]{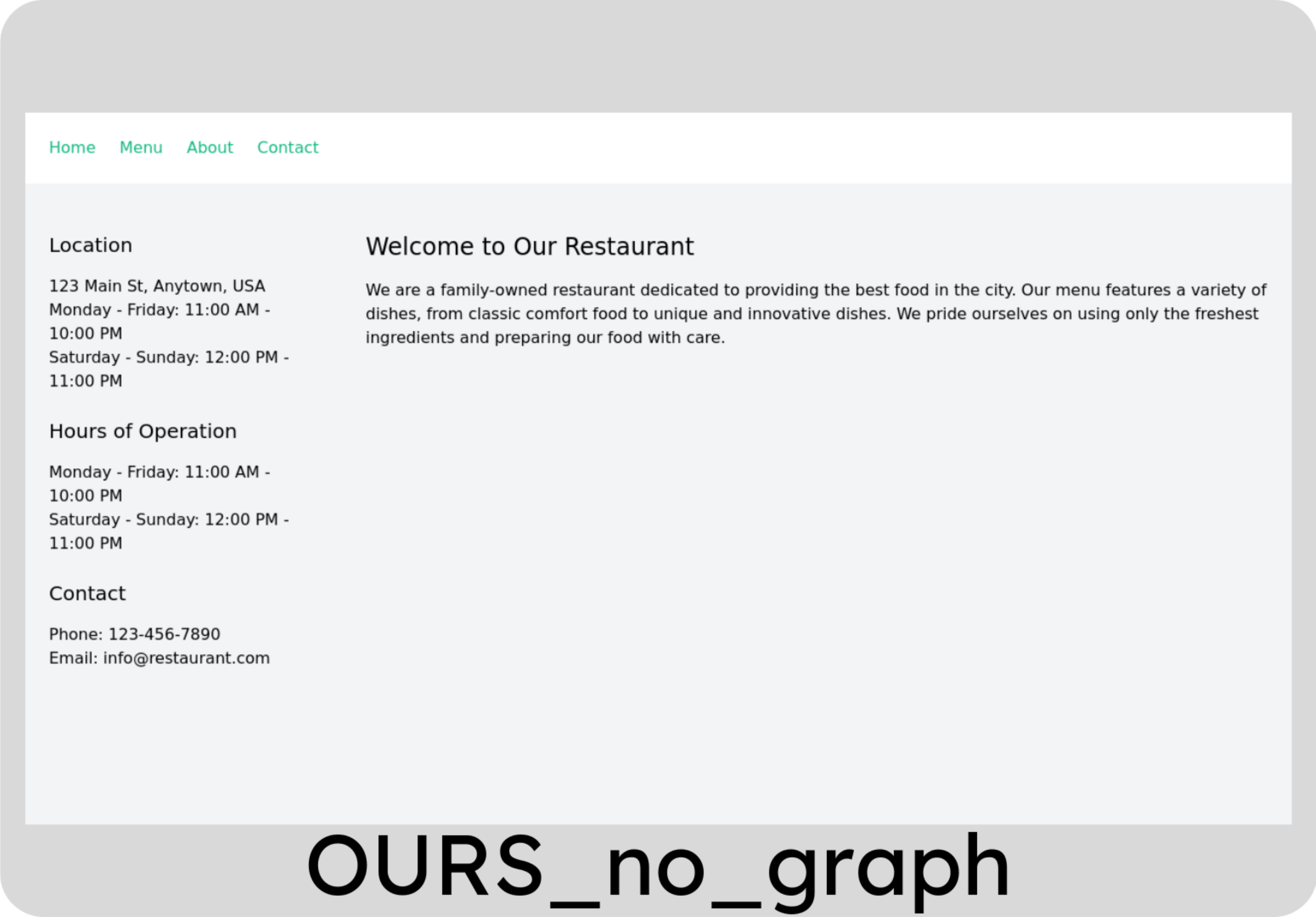} \\

\includegraphics[width=.2\linewidth]{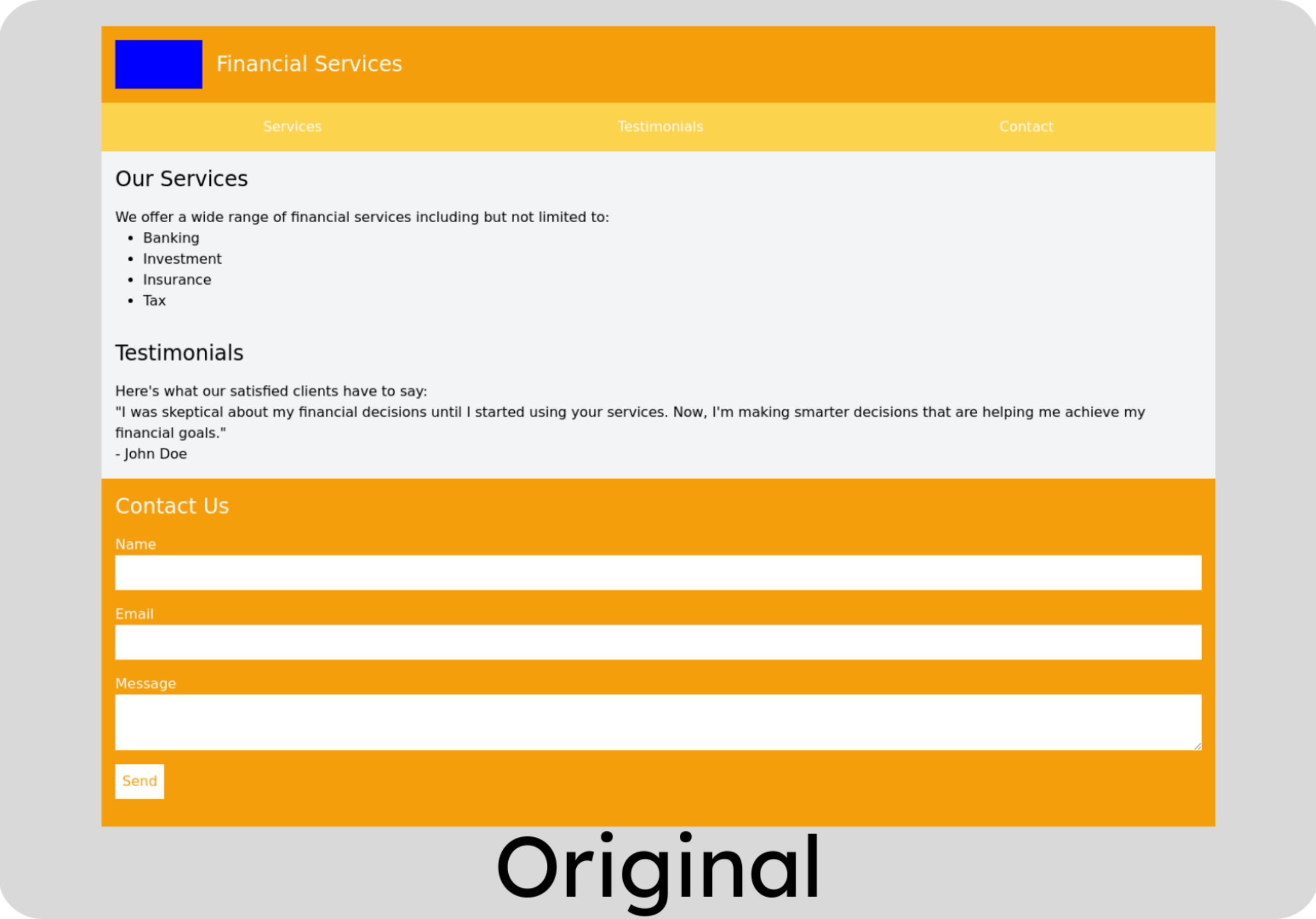} & \includegraphics[width=.2\linewidth]{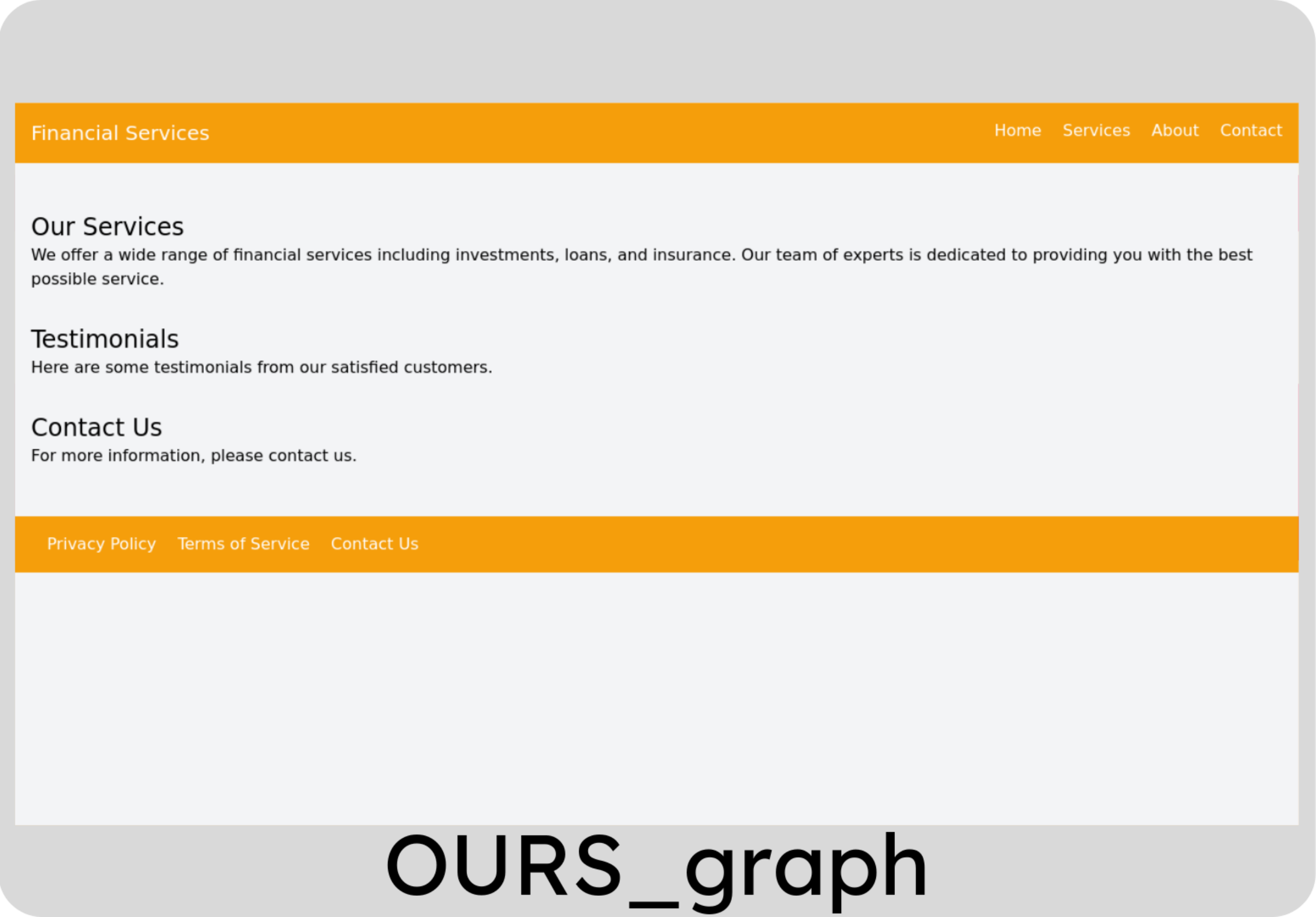} & \includegraphics[width=.2\linewidth]{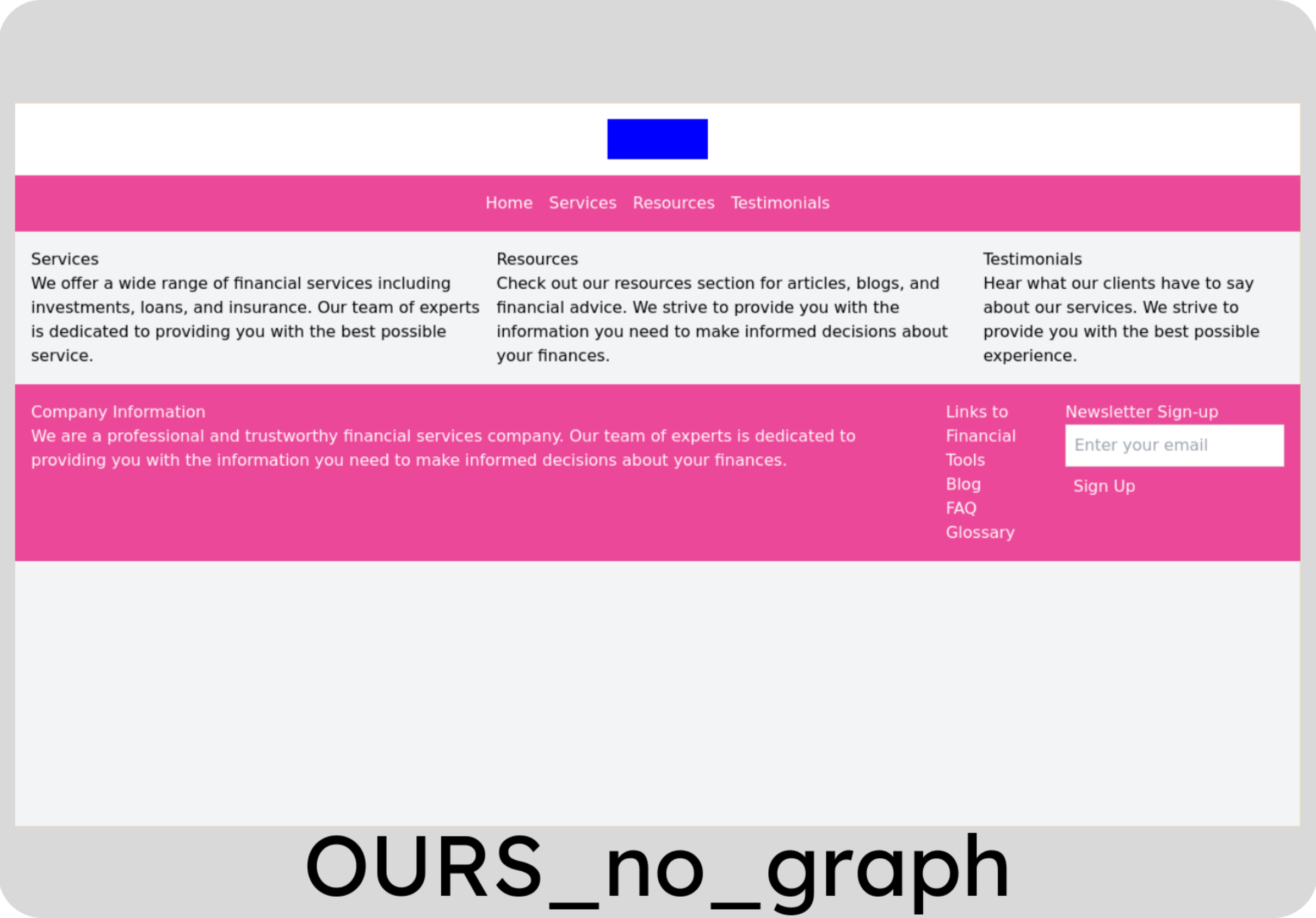} \\

\end{tabular}
}
\end{table}

The performance comparison between OURS-graph, OURS-no-graph, and Gemini prompting demonstrates the clear advantage of using graph-based representations. OURS-graph consistently achieves better structural harmony across webpage components, capturing spatial relationships more accurately and delivering modern, well-aligned designs. It also excels in visual aspects, such as coloring and style. While OURS-no-graph generates functional HTML, it lacks the precision in layout and visual consistency. Gemini prompting performs well in capturing both content and structure but occasionally falls short in maintaining accurate coloring and style. Overall, OURS-graph proves superior in balancing both structure and visual fidelity.

\section{Conclusion} \label{sec:conclusion}
In conclusion, we present a novel approach to solving the Design2Code problem by integrating multimodal information through a graph-based model that captures the spatial and semantic relationships within webpage designs. Our OCR-Segmentation pipeline effectively isolates textual content, allowing for more accurate segmentation of visual elements. The proposed graph-enhanced vision-language model bridges the gap between visual comprehension and code generation, resulting in improved HTML generation in terms of both layout fidelity and content accuracy. Our extensive experiments validate the effectiveness of this approach, offering a significant advancement toward automated code generation from web designs.
\section{Limitation and Future Research} \label{sec:limitation}

One key limitation of our approach is that we do not have enough computing resources to scale up the size of the graph-enhanced vision language models even though the performance of the model with the current model size is promising. Additionally, our proposed graph structure for webpage screenshot does not support dynamic and interactive webpage elements, which occurs regularly in the real-world scenario.

Future research could delve into extending the graph-based representation to model interactive elements. Graphs are well-suited for representing relationships between components, and this can be expanded to capture the temporal and behavioral aspects of dynamic content. For instance, nodes and edges could represent not only static components and their spatial relationships but also interactions, such as hover effects, clicks, and animations. By incorporating state transitions or event-driven behaviors into the graph, the model could generate more comprehensive representations of how a webpage functions, making the framework applicable not only to static HTML but also to dynamic web environments that involve JavaScript or other client-side scripting languages.

\bibliography{main}

\begin{thebibliography}{36}
\providecommand{\natexlab}[1]{#1}
\providecommand{\url}[1]{\texttt{#1}}
\expandafter\ifx\csname urlstyle\endcsname\relax
  \providecommand{\doi}[1]{doi: #1}\else
  \providecommand{\doi}{doi: \begingroup \urlstyle{rm}\Url}\fi

\bibitem[Alayrac et~al.(2022)Alayrac, Donahue, Luc, Miech, Barr, Hasson, Lenc,
  Mensch, Millican, Reynolds, Ring, Rutherford, Cabi, Han, Gong, Samangooei,
  Monteiro, Menick, Borgeaud, Brock, Nematzadeh, Sharifzadeh, Binkowski,
  Barreira, Vinyals, Zisserman, and Simonyan]{alayrac2022flamingo}
Jean-Baptiste Alayrac, Jeff Donahue, Pauline Luc, Antoine Miech, Iain Barr,
  Yana Hasson, Karel Lenc, Arthur Mensch, Katherine Millican, Malcolm Reynolds,
  Roman Ring, Eliza Rutherford, Serkan Cabi, Tengda Han, Zhitao Gong, Sina
  Samangooei, Marianne Monteiro, Jacob Menick, Sebastian Borgeaud, Andrew
  Brock, Aida Nematzadeh, Sahand Sharifzadeh, Mikolaj Binkowski, Ricardo
  Barreira, Oriol Vinyals, Andrew Zisserman, and Karen Simonyan.
\newblock Flamingo: a visual language model for few-shot learning.
\newblock In Alice~H. Oh, Alekh Agarwal, Danielle Belgrave, and Kyunghyun Cho
  (eds.), \emph{Advances in Neural Information Processing Systems}, 2022.
\newblock URL \url{https://openreview.net/forum?id=EbMuimAbPbs}.

\bibitem[Awadalla et~al.(2023)Awadalla, Gao, Gardner, Hessel, Hanafy, Zhu,
  Marathe, Bitton, Gadre, Sagawa, Jitsev, Kornblith, Koh, Ilharco, Wortsman,
  and Schmidt]{awadalla2023openflamingoopensourceframeworktraining}
Anas Awadalla, Irena Gao, Josh Gardner, Jack Hessel, Yusuf Hanafy, Wanrong Zhu,
  Kalyani Marathe, Yonatan Bitton, Samir Gadre, Shiori Sagawa, Jenia Jitsev,
  Simon Kornblith, Pang~Wei Koh, Gabriel Ilharco, Mitchell Wortsman, and Ludwig
  Schmidt.
\newblock Openflamingo: An open-source framework for training large
  autoregressive vision-language models, 2023.
\newblock URL \url{https://arxiv.org/abs/2308.01390}.

\bibitem[Beltramelli(2018)]{10.1145/3220134.3220135}
Tony Beltramelli.
\newblock pix2code: Generating code from a graphical user interface screenshot.
\newblock In \emph{Proceedings of the ACM SIGCHI Symposium on Engineering
  Interactive Computing Systems}, EICS '18, New York, NY, USA, 2018.
  Association for Computing Machinery.
\newblock ISBN 9781450358972.
\newblock \doi{10.1145/3220134.3220135}.
\newblock URL \url{https://doi.org/10.1145/3220134.3220135}.

\bibitem[Bo et~al.(2023)Bo, Luo, and Feng]{GUICG2023}
Cai Bo, Jian Luo, and Zhen Feng.
\newblock A novel code generator for graphical user interfaces.
\newblock \emph{Scientific Reports}, 13, 11 2023.
\newblock \doi{10.1038/s41598-023-46500-6}.

\bibitem[Chen et~al.(2020)Chen, Li, Yu, El~Kholy, Ahmed, Gan, Cheng, and
  Liu]{10.1007/978-3-030-58577-8_7}
Yen-Chun Chen, Linjie Li, Licheng Yu, Ahmed El~Kholy, Faisal Ahmed, Zhe Gan,
  Yu~Cheng, and Jingjing Liu.
\newblock Uniter: Universal image-text representation learning.
\newblock In \emph{Computer Vision – ECCV 2020: 16th European Conference,
  Glasgow, UK, August 23–28, 2020, Proceedings, Part XXX}, pp.\  104–120,
  Berlin, Heidelberg, 2020. Springer-Verlag.
\newblock ISBN 978-3-030-58576-1.
\newblock \doi{10.1007/978-3-030-58577-8_7}.
\newblock URL \url{https://doi.org/10.1007/978-3-030-58577-8_7}.

\bibitem[Devlin et~al.(2019)Devlin, Chang, Lee, and
  Toutanova]{devlin2019bertpretrainingdeepbidirectional}
Jacob Devlin, Ming-Wei Chang, Kenton Lee, and Kristina Toutanova.
\newblock Bert: Pre-training of deep bidirectional transformers for language
  understanding, 2019.
\newblock URL \url{https://arxiv.org/abs/1810.04805}.

\bibitem[Gui et~al.(2024)Gui, Li, Wan, Shi, Zhang, Su, Dong, Zhou, and
  Jiang]{Gui2024VISION2UIAR}
Yi~Gui, Zhen Li, Yao Wan, Yemin Shi, Hongyu Zhang, Yi~Su, Shaoling Dong, Xing
  Zhou, and Wenbin Jiang.
\newblock Vision2ui: A real-world dataset with layout for code generation from
  ui designs.
\newblock \emph{ArXiv}, abs/2404.06369, 2024.
\newblock URL \url{https://api.semanticscholar.org/CorpusID:269010048}.

\bibitem[Gui et~al.(2025)Gui, Wan, Li, Zhang, Chen, Zhang, Su, Chen, Zhou,
  Jiang, and Zhang]{gui2025uicopilot}
Yi~Gui, Yao Wan, Zhen Li, Zhongyi Zhang, Dongping Chen, Hongyu Zhang, Yi~Su,
  Bohua Chen, Xing Zhou, Wenbin Jiang, and Xiangliang Zhang.
\newblock {UIC}opilot: Automating {UI} synthesis via hierarchical code
  generation from webpage designs.
\newblock In \emph{THE WEB CONFERENCE 2025}, 2025.
\newblock URL \url{https://openreview.net/forum?id=faMbH0wkye}.

\bibitem[Hakimov \& Schlangen(2023)Hakimov and
  Schlangen]{hakimov-schlangen-2023-images}
Sherzod Hakimov and David Schlangen.
\newblock Images in language space: Exploring the suitability of large language
  models for vision {\&} language tasks.
\newblock In Anna Rogers, Jordan Boyd-Graber, and Naoaki Okazaki (eds.),
  \emph{Findings of the Association for Computational Linguistics: ACL 2023},
  pp.\  14196--14210, Toronto, Canada, July 2023. Association for Computational
  Linguistics.
\newblock \doi{10.18653/v1/2023.findings-acl.894}.
\newblock URL \url{https://aclanthology.org/2023.findings-acl.894}.

\bibitem[Iscen et~al.(2024)Iscen, Caron, Fathi, and
  Schmid]{iscen2024retrievalenhancedcontrastivevisiontextmodels}
Ahmet Iscen, Mathilde Caron, Alireza Fathi, and Cordelia Schmid.
\newblock Retrieval-enhanced contrastive vision-text models, 2024.
\newblock URL \url{https://arxiv.org/abs/2306.07196}.

\bibitem[Jaegle et~al.(2021)Jaegle, Gimeno, Brock, Vinyals, Zisserman, and
  Carreira]{perceiver}
Andrew Jaegle, Felix Gimeno, Andy Brock, Oriol Vinyals, Andrew Zisserman, and
  Joao Carreira.
\newblock Perceiver: General perception with iterative attention.
\newblock In Marina Meila and Tong Zhang (eds.), \emph{Proceedings of the 38th
  International Conference on Machine Learning}, volume 139 of
  \emph{Proceedings of Machine Learning Research}, pp.\  4651--4664. PMLR,
  18--24 Jul 2021.
\newblock URL \url{https://proceedings.mlr.press/v139/jaegle21a.html}.

\bibitem[Jiang et~al.(2024)Jiang, Zhou, Garg, and Oulasvirta]{graph4gui}
Yue Jiang, Changkong Zhou, Vikas Garg, and Antti Oulasvirta.
\newblock Graph4gui: Graph neural networks for representing graphical user
  interfaces.
\newblock In \emph{Proceedings of the 2024 CHI Conference on Human Factors in
  Computing Systems}, CHI '24, New York, NY, USA, 2024. Association for
  Computing Machinery.
\newblock ISBN 9798400703300.
\newblock \doi{10.1145/3613904.3642822}.
\newblock URL \url{https://doi.org/10.1145/3613904.3642822}.

\bibitem[Kipf \& Welling(2017)Kipf and Welling]{kipf2017semisupervised}
Thomas~N. Kipf and Max Welling.
\newblock Semi-supervised classification with graph convolutional networks.
\newblock In \emph{International Conference on Learning Representations}, 2017.
\newblock URL \url{https://openreview.net/forum?id=SJU4ayYgl}.

\bibitem[Lauren\c{c}on et~al.(2023)Lauren\c{c}on, Saulnier, Tronchon, Bekman,
  Singh, Lozhkov, Wang, Karamcheti, Rush, Kiela, Cord, and
  Sanh]{NEURIPS2023_e2cfb719}
Hugo Lauren\c{c}on, Lucile Saulnier, Leo Tronchon, Stas Bekman, Amanpreet
  Singh, Anton Lozhkov, Thomas Wang, Siddharth Karamcheti, Alexander Rush,
  Douwe Kiela, Matthieu Cord, and Victor Sanh.
\newblock Obelics: An open web-scale filtered dataset of interleaved image-text
  documents.
\newblock In A.~Oh, T.~Naumann, A.~Globerson, K.~Saenko, M.~Hardt, and
  S.~Levine (eds.), \emph{Advances in Neural Information Processing Systems},
  volume~36, pp.\  71683--71702. Curran Associates, Inc., 2023.
\newblock URL
  \url{https://proceedings.neurips.cc/paper_files/paper/2023/file/e2cfb719f58585f779d0a4f9f07bd618-Paper-Datasets_and_Benchmarks.pdf}.

\bibitem[Laurençon et~al.(2024)Laurençon, Tronchon, and Sanh]{Hugo2024}
Hugo Laurençon, Léo Tronchon, and Victor Sanh.
\newblock Unlocking the conversion of web screenshots into html code with the
  websight dataset, 2024.
\newblock URL \url{https://arxiv.org/abs/2403.09029}.

\bibitem[Li et~al.(2021)Li, Selvaraju, Gotmare, Joty, Xiong, and
  Hoi]{NEURIPS2021_50525975}
Junnan Li, Ramprasaath Selvaraju, Akhilesh Gotmare, Shafiq Joty, Caiming Xiong,
  and Steven Chu~Hong Hoi.
\newblock Align before fuse: Vision and language representation learning with
  momentum distillation.
\newblock In M.~Ranzato, A.~Beygelzimer, Y.~Dauphin, P.S. Liang, and J.~Wortman
  Vaughan (eds.), \emph{Advances in Neural Information Processing Systems},
  volume~34, pp.\  9694--9705. Curran Associates, Inc., 2021.
\newblock URL
  \url{https://proceedings.neurips.cc/paper_files/paper/2021/file/505259756244493872b7709a8a01b536-Paper.pdf}.

\bibitem[Li et~al.(2022)Li, Li, Xiong, and Hoi]{pmlr-v162-li22n}
Junnan Li, Dongxu Li, Caiming Xiong, and Steven Hoi.
\newblock {BLIP}: Bootstrapping language-image pre-training for unified
  vision-language understanding and generation.
\newblock In Kamalika Chaudhuri, Stefanie Jegelka, Le~Song, Csaba Szepesvari,
  Gang Niu, and Sivan Sabato (eds.), \emph{Proceedings of the 39th
  International Conference on Machine Learning}, volume 162 of
  \emph{Proceedings of Machine Learning Research}, pp.\  12888--12900. PMLR,
  17--23 Jul 2022.
\newblock URL \url{https://proceedings.mlr.press/v162/li22n.html}.

\bibitem[Li et~al.(2019)Li, Yatskar, Yin, Hsieh, and
  Chang]{li2019visualbertsimpleperformantbaseline}
Liunian~Harold Li, Mark Yatskar, Da~Yin, Cho-Jui Hsieh, and Kai-Wei Chang.
\newblock Visualbert: A simple and performant baseline for vision and language,
  2019.
\newblock URL \url{https://arxiv.org/abs/1908.03557}.

\bibitem[Lin et~al.(2024)Lin, Yin, Ping, Molchanov, Shoeybi, and
  Han]{Lin_2024_CVPR}
Ji~Lin, Hongxu Yin, Wei Ping, Pavlo Molchanov, Mohammad Shoeybi, and Song Han.
\newblock Vila: On pre-training for visual language models.
\newblock In \emph{Proceedings of the IEEE/CVF Conference on Computer Vision
  and Pattern Recognition (CVPR)}, pp.\  26689--26699, June 2024.

\bibitem[Liu et~al.(2019)Liu, Nickel, and Kiela]{NEURIPS2019_103303dd}
Qi~Liu, Maximilian Nickel, and Douwe Kiela.
\newblock Hyperbolic graph neural networks.
\newblock In H.~Wallach, H.~Larochelle, A.~Beygelzimer, F.~d\textquotesingle
  Alch\'{e}-Buc, E.~Fox, and R.~Garnett (eds.), \emph{Advances in Neural
  Information Processing Systems}, volume~32. Curran Associates, Inc., 2019.
\newblock URL
  \url{https://proceedings.neurips.cc/paper_files/paper/2019/file/103303dd56a731e377d01f6a37badae3-Paper.pdf}.

\bibitem[Lyu et~al.(2024)Lyu, Ray, Roychoudhury, Tan, and
  Thongtanunam]{lyu2024automaticprogramminglargelanguage}
Michael~R. Lyu, Baishakhi Ray, Abhik Roychoudhury, Shin~Hwei Tan, and Patanamon
  Thongtanunam.
\newblock Automatic programming: Large language models and beyond, 2024.
\newblock URL \url{https://arxiv.org/abs/2405.02213}.

\bibitem[Nguyen \& Csallner(2015)Nguyen and Csallner]{Nguyen2015ReverseEM}
Tuan~Anh Nguyen and Christoph Csallner.
\newblock Reverse engineering mobile application user interfaces with remaui
  (t).
\newblock \emph{2015 30th IEEE/ACM International Conference on Automated
  Software Engineering (ASE)}, pp.\  248--259, 2015.
\newblock URL \url{https://api.semanticscholar.org/CorpusID:7499368}.

\bibitem[Radford et~al.(2021)Radford, Kim, Hallacy, Ramesh, Goh, Agarwal,
  Sastry, Askell, Mishkin, Clark, Krueger, and Sutskever]{pmlr-v139-radford21a}
Alec Radford, Jong~Wook Kim, Chris Hallacy, Aditya Ramesh, Gabriel Goh,
  Sandhini Agarwal, Girish Sastry, Amanda Askell, Pamela Mishkin, Jack Clark,
  Gretchen Krueger, and Ilya Sutskever.
\newblock Learning transferable visual models from natural language
  supervision.
\newblock In Marina Meila and Tong Zhang (eds.), \emph{Proceedings of the 38th
  International Conference on Machine Learning}, volume 139 of
  \emph{Proceedings of Machine Learning Research}, pp.\  8748--8763. PMLR,
  18--24 Jul 2021.
\newblock URL \url{https://proceedings.mlr.press/v139/radford21a.html}.

\bibitem[Ramp\'{a}\v{s}ek et~al.(2022)Ramp\'{a}\v{s}ek, Galkin, Dwivedi, Luu,
  Wolf, and Beaini]{NEURIPS2022_5d4834a1}
Ladislav Ramp\'{a}\v{s}ek, Michael Galkin, Vijay~Prakash Dwivedi, Anh~Tuan Luu,
  Guy Wolf, and Dominique Beaini.
\newblock Recipe for a general, powerful, scalable graph transformer.
\newblock In S.~Koyejo, S.~Mohamed, A.~Agarwal, D.~Belgrave, K.~Cho, and A.~Oh
  (eds.), \emph{Advances in Neural Information Processing Systems}, volume~35,
  pp.\  14501--14515. Curran Associates, Inc., 2022.
\newblock URL
  \url{https://proceedings.neurips.cc/paper_files/paper/2022/file/5d4834a159f1547b267a05a4e2b7cf5e-Paper-Conference.pdf}.

\bibitem[Robinson(2019)]{robinson2019sketch2codegeneratingwebsitepaper}
Alex Robinson.
\newblock Sketch2code: Generating a website from a paper mockup, 2019.
\newblock URL \url{https://arxiv.org/abs/1905.13750}.

\bibitem[Ruiz et~al.(2020)Ruiz, Gama, and Ribeiro]{9239975}
Luana Ruiz, Fernando Gama, and Alejandro Ribeiro.
\newblock Gated graph recurrent neural networks.
\newblock \emph{IEEE Transactions on Signal Processing}, 68:\penalty0
  6303--6318, 2020.
\newblock \doi{10.1109/TSP.2020.3033962}.

\bibitem[ShantamVijayputra(2022)]{shantamvijayputra-2022}
ShantamVijayputra.
\newblock {Sketch2Code}, 12 2022.
\newblock URL \url{https://www.kaggle.com/datasets/vshantam/sketch2code}.

\bibitem[Si et~al.(2024)Si, Zhang, Yang, Liu, and
  Yang]{si2024design2codefarautomatingfrontend}
Chenglei Si, Yanzhe Zhang, Zhengyuan Yang, Ruibo Liu, and Diyi Yang.
\newblock Design2code: How far are we from automating front-end engineering?,
  2024.
\newblock URL \url{https://arxiv.org/abs/2403.03163}.

\bibitem[Soselia et~al.(2023)Soselia, Saifullah, and
  Zhou]{soselia2023learninguitocodereversegenerator}
Davit Soselia, Khalid Saifullah, and Tianyi Zhou.
\newblock Learning ui-to-code reverse generator using visual critic without
  rendering, 2023.
\newblock URL \url{https://arxiv.org/abs/2305.14637}.

\bibitem[Sun et~al.(2024)Sun, Chen, Xu, Cheng, Ma, Yin, Wang, Han, Zhu, Yuan,
  Guo, Qiu, Yin, Li, Yuan, Kong, Li, and
  Wu]{sun2024surveyneuralcodeintelligence}
Qiushi Sun, Zhirui Chen, Fangzhi Xu, Kanzhi Cheng, Chang Ma, Zhangyue Yin,
  Jianing Wang, Chengcheng Han, Renyu Zhu, Shuai Yuan, Qipeng Guo, Xipeng Qiu,
  Pengcheng Yin, Xiaoli Li, Fei Yuan, Lingpeng Kong, Xiang Li, and Zhiyong Wu.
\newblock A survey of neural code intelligence: Paradigms, advances and beyond,
  2024.
\newblock URL \url{https://arxiv.org/abs/2403.14734}.

\bibitem[Tan \& Bansal(2019)Tan and Bansal]{tan-bansal-2019-lxmert}
Hao Tan and Mohit Bansal.
\newblock {LXMERT}: Learning cross-modality encoder representations from
  transformers.
\newblock In Kentaro Inui, Jing Jiang, Vincent Ng, and Xiaojun Wan (eds.),
  \emph{Proceedings of the 2019 Conference on Empirical Methods in Natural
  Language Processing and the 9th International Joint Conference on Natural
  Language Processing (EMNLP-IJCNLP)}, pp.\  5100--5111, Hong Kong, China,
  November 2019. Association for Computational Linguistics.
\newblock \doi{10.18653/v1/D19-1514}.
\newblock URL \url{https://aclanthology.org/D19-1514}.

\bibitem[Wang et~al.(2024)Wang, Lv, Yu, Hong, Qi, Wang, Ji, Yang, Zhao, Song,
  Xu, Xu, Li, Dong, Ding, and Tang]{wang2024cogvlmvisualexpertpretrained}
Weihan Wang, Qingsong Lv, Wenmeng Yu, Wenyi Hong, Ji~Qi, Yan Wang, Junhui Ji,
  Zhuoyi Yang, Lei Zhao, Xixuan Song, Jiazheng Xu, Bin Xu, Juanzi Li, Yuxiao
  Dong, Ming Ding, and Jie Tang.
\newblock Cogvlm: Visual expert for pretrained language models, 2024.
\newblock URL \url{https://arxiv.org/abs/2311.03079}.

\bibitem[Wu et~al.(2023)Wu, Wang, Shen, Peng, Nichols, and
  Bigham]{10.1145/3544548.3581158}
Jason Wu, Siyan Wang, Siman Shen, Yi-Hao Peng, Jeffrey Nichols, and Jeffrey~P
  Bigham.
\newblock Webui: A dataset for enhancing visual ui understanding with web
  semantics.
\newblock In \emph{Proceedings of the 2023 CHI Conference on Human Factors in
  Computing Systems}, CHI '23, New York, NY, USA, 2023. Association for
  Computing Machinery.
\newblock ISBN 9781450394215.
\newblock \doi{10.1145/3544548.3581158}.
\newblock URL \url{https://doi.org/10.1145/3544548.3581158}.

\bibitem[Yun et~al.(2024)Yun, Lin, Thushara, Bhat, Wang, Jiang, Deng, Wang,
  Tao, Li, Li, Nakov, Baldwin, Liu, Xing, Liang, and Shen]{Yun2024Web2CodeAL}
Sukmin Yun, Haokun Lin, Rusiru Thushara, Mohammad~Qazim Bhat, Yongxin Wang,
  Zutao Jiang, Mingkai Deng, Jinhong Wang, Tianhua Tao, Junbo Li, Haonan Li,
  Preslav Nakov, Timothy Baldwin, Zhengzhong Liu, Eric~P. Xing, Xiaodan Liang,
  and Zhiqiang Shen.
\newblock Web2code: A large-scale webpage-to-code dataset and evaluation
  framework for multimodal llms.
\newblock \emph{ArXiv}, abs/2406.20098, 2024.
\newblock URL \url{https://api.semanticscholar.org/CorpusID:270845897}.

\bibitem[Zan et~al.(2023)Zan, Chen, Zhang, Lu, Wu, Guan, Wang, and
  Lou]{zan2023largelanguagemodelsmeet}
Daoguang Zan, Bei Chen, Fengji Zhang, Dianjie Lu, Bingchao Wu, Bei Guan, Yongji
  Wang, and Jian-Guang Lou.
\newblock Large language models meet nl2code: A survey, 2023.
\newblock URL \url{https://arxiv.org/abs/2212.09420}.

\bibitem[Zhou et~al.(2024)Zhou, Zhao, Hou, Sun, Chen, and Wang]{declareUI}
Ting Zhou, Yanjie Zhao, Xinyi Hou, Xiaoyu Sun, Kai Chen, and Haoyu Wang.
\newblock Bridging design and development with automated declarative ui code
  generation, 2024.
\newblock URL \url{https://arxiv.org/abs/2409.11667}.

\end{thebibliography}
\bibliographystyle{iclr2025_conference}


\end{document}